\documentclass{article}

\usepackage{PRIMEarxiv}

\usepackage[utf8]{inputenc} 
\usepackage[T1]{fontenc}    
\usepackage{hyperref}       
\usepackage{url}            
\usepackage{booktabs}       
\usepackage{amsfonts}       
\usepackage{nicefrac}       
\usepackage{microtype}      
\usepackage{lipsum}
\usepackage{fancyhdr}       
\usepackage{graphicx}       
\usepackage{times}
\usepackage{latexsym}
\usepackage{amsmath}
\usepackage{xspace}
\usepackage{multirow}
\usepackage{graphicx}
\usepackage{caption}
\usepackage{subcaption}
\usepackage{hyperref}
\usepackage{tabularx}
\graphicspath{{media/}}     

\title{\textit{Coherence and Diversity through Noise}: Self-Supervised Paraphrase Generation via Structure-Aware Denoising
}

\author{
  Rishabh Gupta, Venktesh V., Mukesh Mohania, Vikram Goyal \\
  Department of CSE, IIIT-Delhi \\
  \texttt{\{rishabh19089, venkteshv, mukesh, vikram\}@iiitd.ac.in} \\
}

\newcommand{\p}{\Psi}
\newcommand{\doc}[1]{$d_{#1}$}
\newcommand{\setb}[2]{\{#1^{1}, #1^{2} \dots #1^{#2}\}}
\newcommand{\tupleb}[2]{(#1_{1}, #1_{2} \dots #1_{#2})}
\newcommand{\tuplec}[3]{(#1_{1}^{#3}, #1_{2}^{#3} \dots #1_{#2}^{#3})}
\newcommand{\simi}[2]{\sigma_{x^{#1}, y^{#2}}}
\newcommand{\dive}[2]{\delta_{x^{#1}, y^{#2}}}
\newcommand{\nume}[2]{\nu_{x^{#1}, y^{#2}}}
\newcommand{\ove}[2]{\lambda_{x^{#1}, y^{#2}}}

\newcommand{\dn}{\Delta}
\newcommand{\corp}[1]{\{x^{1}, x^{2} \dots x^{#1}\}}

\newcommand{\tna}[1]{q^t_{\pi_#1}}
\newcommand{\tni}[1]{q^t_{#1}}

\newcommand{\inna}[1]{q^i_{\pi_#1}}
\newcommand{\inni}[1]{q^i_#1}
\newcommand{\paco}[1]{\{(x^1, \hat{y}^1), (x^2, \hat{y}^2) \dots (x^{#1}, \hat{y}^{#1})\}}
\newcommand{\pacop}[1]{\{(x^1, y^1, p^1), (x^2, y^2, p^2) \dots (x^{#1}, y^{#1}, p^#1)\}}
\newcommand{\smc}[2]{\hat{x}^{#1} \sim q_{\pi_{#2}}(\hat{x}^{#1}|x^{#1})}

\newcommand{\exit}[1]{\hat{e}^t_#1}
\newcommand{\exii}[1]{\hat{e}^i_#1}
\newcommand{\exct}[1]{\hat{e}^t_{\pi_#1}}
\newcommand{\exci}[1]{\hat{e}^i_{\pi_#1}}
\newcommand{\pan}{\texttt{PAN}\xspace}
\newcommand{\model}{\texttt{SCANING}\xspace}
\newcommand{\modela}{\texttt{STEAD}\xspace}
\newcommand{\modelb}{\texttt{PAP}\xspace}

\begin{document}
\maketitle

\begin{abstract}
In this paper, we propose \model, an unsupervised framework for paraphrasing via controlled noise injection. We focus on the novel task of paraphrasing algebraic word problems having practical applications in online pedagogy as a means to reduce plagiarism as well as ensure understanding on the part of the student instead of rote memorization. This task is more complex than paraphrasing general-domain corpora due to the difficulty in preserving critical information for solution consistency of the paraphrased word problem, managing the increased length of the text and ensuring diversity in the generated paraphrase. Existing approaches fail to demonstrate adequate performance on at least one, if not all, of these facets, necessitating the need for a more comprehensive solution. To this end, we model the noising search space as a composition of contextual and syntactic aspects and sample noising functions consisting of either one or both aspects. This allows for learning a denoising function that operates over both aspects and produces semantically equivalent and syntactically diverse outputs through grounded noise injection. The denoising function serves as a foundation for learning a paraphrasing function which operates solely in the input-paraphrase space without carrying any direct dependency on noise. We demonstrate \model considerably improves performance in terms of both semantic preservation and producing diverse paraphrases through extensive automated and manual evaluation across $4$ datasets.
\end{abstract}


\section{Introduction}
\begin{figure}[h]
    \centering
    \includegraphics[width=\textwidth]{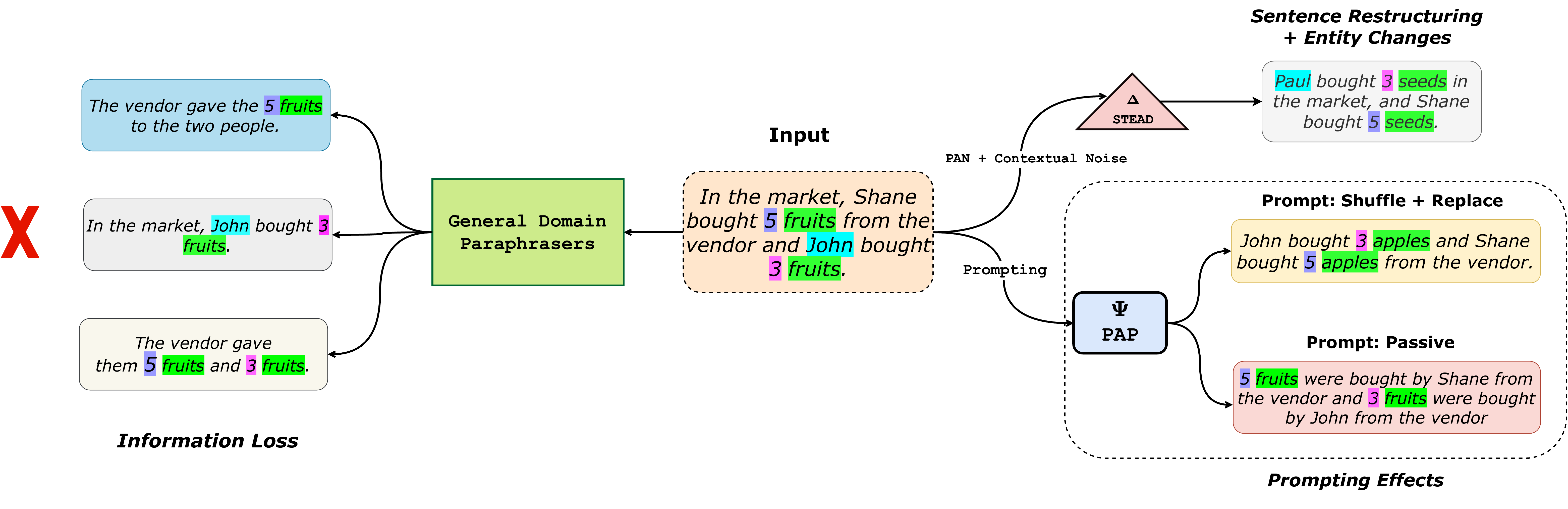}
    \caption{Overview of the outputs of our models (\modela and \modelb) as well as paraphrasers trained on general-domain corpora (for instance: Pegasus fine-tuned on PAWS). The general domain paraphrasers frequently lose critical information, rendering the paraphrased problem unsolvable \cite{https://doi.org/10.48550/arxiv.2206.08263}. Our models generate semantically coherent yet syntactically diverse outputs as shown. In this instance, the denoiser \modela generates outputs which display sentence restructuring as well as entity changes through the application of different types of noise. Similarly, given different prompts, \modelb generates distinct and diverse outputs, including a passivized version.}
    \label{fig:eg}
    \label{fig:prompt}
\end{figure}


Paraphrasing textual content is a vital task in the field of NLP due to its wide-scale applications for tasks such as question answering \cite{buck2017ask}, summarization \cite{cao2017joint}, amongst others. We study a use case for paraphrasing in online pedagogical settings. Online educational platforms conduct assessments to evaluate the users. However, the repetition of question phrasing can lead to memorization and the use of unethical methods, necessitating the creation of varied questions to alter the form. Generating multiple variations of the same question and distributing them among different students helps to maintain consistency in difficulty level and prevent the reliance on memorization. Our motivation is also grounded in learning theory\footnote{\url{https://cutt.ly/MWqHsN8}}. The underlying hypothesis posits that by converting the linguistic information into one's own words while maintaining the essence, students can enhance their comprehension of the problem to a greater extent as compared to mere rote memorization. The theory stipulates that rewording the information while preserving the significance facilitates students in developing their own reasoning behind the sequence of steps to reach a solution. Our endeavor is founded upon the same premise, seeking to restate the problem without altering the resolution, thus promoting diversity in the phrasing of the problem and encouraging students to form their own interpretations from the rephrased versions. Furthermore, this approach facilitates assessment for educators while retaining the ability to generate a diverse range of questions.

To this end, we focus on algebraic word problems (AWPs) and propose the novel task of paraphrasing AWPs. We define the paraphrasing of AWPs as \textit{changing surface form of the original problem while preserving the underlying equation and solution.} Paraphrasing AWPs is more complex than paraphrasing general-domain text, as AWPs require preserving crucial aspects like numbers and units to ensure solution consistency.
Further, word problems usually consist of multiple sentences and thus require coherence and consistency in the generated paraphrase over an extended context. Another facet crucial to our task is the diversity of the generated paraphrase due to the inherent characteristics of AWPs, wherein a change in entities along with structure yields valid paraphrases if the underlying equation is unperturbed. Although existing approaches \cite{prakash-etal-2016-neural, liu-etal-2020-unsupervised, uPA} demonstrate competent results on general-domain corpora, they tend to hallucinate or omit key entities when applied to the task of paraphrasing AWPs, thus rendering questions unsolvable (c.f. Fig \ref{fig:eg}). 

We identify two significant components of paraphrasing algebraic word problems: changing the entities (contextual changes) and changing the structure of the sentences (syntactic changes). We rigorously study and design noising functions that take these aspects into consideration and \textit{exploit} them to learn a denoiser $\dn$ that operates over both aspects to change the surface form while retaining semantic equivalence through grounded knowledge fusion. $\dn$ is then used to generate a parallel corpus to serve as a base for learning a paraphrasing function, $\p$. $\p$ is not directly dependent on the noise as it learns to map from input to paraphrase (Fig \ref{fig:pipeline}), and any supervised paraphrasing method can serve to parameterize $\p$ as our method is model-agnostic. We refer to the parameterized model of $\dn$ as \modela (STructurE Aware Denoiser) and the parameterized model of $\p$ as \modelb (Prompt Aware Paraphraser). Note that this terminology is used interchangeably throughout the paper.

To summarize, the core contributions of our work include: 
\begin{itemize}
    \item \textbf{Novel Task} -- We propose the new task of paraphrasing algebraic word problems.
    \item \textbf{Novel Noise Injection} -- We define and analyze the combinations of multiple novel noising functions by modelling the search space as a composition of syntactical and contextual variations. We introduce the concept of \pan (\textit{Pseudo-Adversarial Noising}) to dupe the denoiser into making syntactic changes via grounding on linguistic regularities. We present a systematic overview of the induced ability from the noising functions and the reasons for their selection.
    \item \textbf{Novel Framework} -- We design \model, a novel prompt-aware two-stage pipeline consisting of a denoiser, \modela and a paraphraser, \modelb to generate semantically rich and syntactically diverse paraphrases (even over long contexts) without the availability of reference paraphrases.
    \item {\bf Evaluation} -- We conduct conduct extensive automated and human evaluation to assess the efficacy of our approach. For automated evaluation, we define new metrics (due to the new task) to quantify the soundness of the generated paraphrases. Both automated and manual results demonstrate that \model significantly outperforms all baselines.
    \item {\bf Analysis} -- We supply comprehensive analysis (both quantitative and qualitative) to gain an insight into the working components of the proposed framework and identify its limitations.
\end{itemize}
We release the anonymized version of our source code and data with instructions for reproducing the results at: \\ \url{https://anonymous.4open.science/r/SCANInG/}.


\section{Related Work}
Paraphrasing of general domain corpus is a well-studied problem and can be categorized into supervised and unsupervised approaches. Earlier approaches \cite{ellsworth-janin-2007-mutaphrase, narayan-etal-2016-paraphrase} leveraged syntactic features or statistical machine translation approaches \cite{quirk-etal-2004-monolingual, dolan-etal-2004-unsupervised} for paraphrasing . The paraphrasing task has also been modelled as a sequence generation task using encoder-decoder architectures \cite{prakash-etal-2016-neural}. More recently, transformers \cite{wang_transformer, goyal-durrett-2020-neural} are widely used for paraphrase generation but traditionally require labelled paraphrases for training which requires a time-consuming annotation process.

Unsupervised approaches like Variational AutoEncoders (VAEs) aid in text generation by directly sampling from the latent space \cite{bowman-etal-2016-generating, zhang-etal-2019-syntax-infused}. However, the sampling is stochastic and controlled generation becomes difficult during decoding resulting in incorrect sentences. To resolve this, several approaches like Unsupervised Data Augmentation (UDA) and SSMBA have proposed data augmentation operators for generating self-supervised corpus for paraphrase generation. UDA leverages operators like backtranslation and word replacement to generate new samples. The authors of SSMBA \cite{ng-etal-2020-ssmba} propose a manifold-based augmentation method where the inputs are re-projected to a different point in the manifold by sampling from latent space. Alternate approaches have been proposed where heuristic search-based methods like simulated annealing are leveraged \cite{liu-etal-2020-unsupervised} to search over edit operators. Since the generated samples are not diverse enough in the above approaches, Rotom \cite{rotom} proposed a set of operators for perturbation to train a generative model. Similarly, GPT-2 has been employed to generate augmented samples \cite{uPA} through simple corruption and reconstruction. There have also been attempts for controlling the paraphrase generation like SynPG \cite{huang-chang-2021-generating} by disentangling the syntactic and semantic components in the embeddings space. However, these approaches still fail to generate diverse and semantically equivalent paraphrases for AWPs.

\section{Methodology}

\begin{figure}[h]
    \centering
    \includegraphics[width=0.45\textwidth]{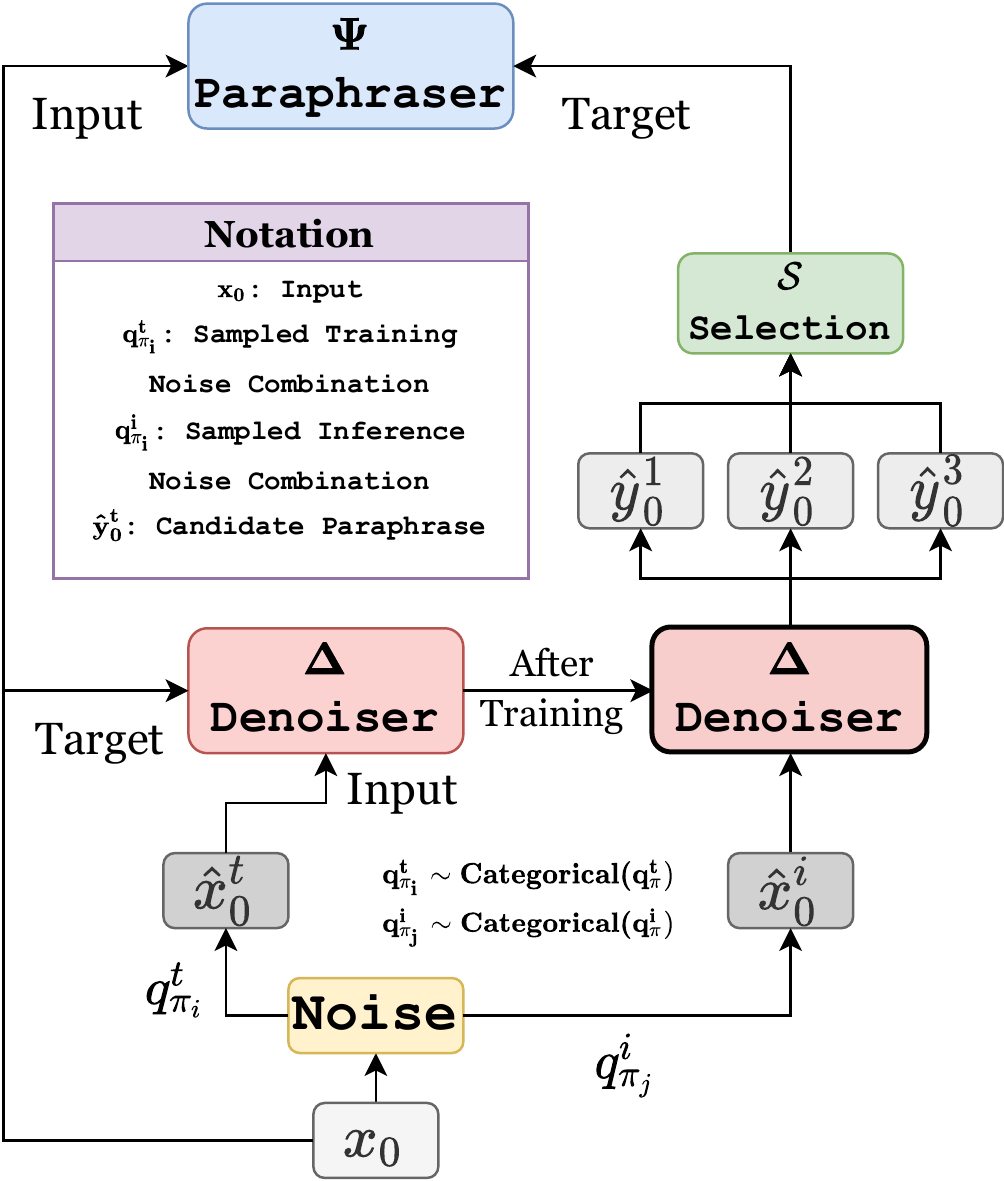}
    \caption{The pipeline of our proposed framework, \model. Given an input sample $x_0$, we sample a noise combination from the training bank to yield $\hat{x}_0^t$, which is given as the input for training the denoiser (\modela) with the original sample $x_0$ as the reconstructive target. After training, a distinct noise combination is sampled from the inference bank, which is given to \modela to generate multiple paraphrases $\hat{y}_0^1, \hat{y}_0^2, \dots \hat{y}_0^l$. A selection mechanism ($QF-PCF$ + $PAMMR$) is applied to the generated paraphrases, and the selected paraphrases are given as the target to train the paraphraser \modelb, with $x_0$ as the input. This removes any direct dependency on the noising combination due to the existence of a direct input $\mapsto$ paraphrase mapping for \modelb.}
    \label{fig:pipeline}
\end{figure}

Due to the lack of annotated corpora for paraphrasing AWPs, we devise a self-supervised method for paraphrasing. Let the original corpus be $D=\corp{n}$ where the document $x^i$ consisting of $k$ tokens is represented as $\tuplec{x}{k}{i}$. We aim to generate a set of valid paraphrases $Y = \setb{y}{n}$ without using any labelled data. We adopt a two-stage approach to accomplish this, wherein in the first stage, we learn $\dn$ and then use it to generate a weakly-supervised parallel corpus $P=\paco{n}$ for learning a paraphrasing function $\p$ in the second stage (Figure \ref{fig:pipeline}). $\p$ directly operates over the input-paraphrase space without carrying any direct dependency on the noise and allows us to generate a paraphrase $y^i$ from the input $x^i$ directly as $y^i \sim p_{\p}(y^i|x^i)$. $\dn$ and $\p$ are both parameterized by separate transformer-based neural networks \cite{vaswani2017attention}.

\subsection{Noising Functions}
Let us denote the space of noising functions by $q$; instead of applying them individually to noise the input,\textit{we sample a combination} of these noising functions from a defined categorical distribution and apply them sequentially. This is done to encourage $\dn$ to learn to denoise multiple noises at once to increase its generalization capability to adapt to drift in the noising space. We denote the $j^{th}$ individual noising function as $q_j$, the $j^{th}$ combination as $q_{\pi_j}$ and the $i^{th}$ corrupted sample as $\hat{x}^{i}$, where $\smc{i}{a}$ for some probabilistic corruption combination $q_{\pi_a}$. We refer to syntactic noise as \textit{s-noise}, contextual noise as \textit{c-noise} and contextual+syntactic noise as \textit{cs-noise} for brevity and ease of understanding.

We enumerate the individual training and inference noising functions in the next two sections. We'll use the following running example $e$ for better explainability and denote the output of the $j^{th}$ noising function and combination as $\exit{j} \sim q^t_j(\hat{e} | e)$ and $\exct{j}  \sim q^t_{\pi_j}(\hat{e} | e)$ respectively. \\
$\mathbf{e}$: \textit{Steve rode his car for 5 miles on the way home.}


\subsection{Training Noise}
As noted above, we model the noising space as a composition of both contextual and syntactic aspects. This view is taken with respect to the changes induced through the denoiser ($\dn$) by the noising function. For instance, the aim of an \textit{s-noise} function is to induce syntactic variations in the output of $\dn$, while the aim of a \textit{c-noise} function is to induce entity changes. We define $7$ distinct individual noising functions $q^t_i$, and $10$ combinations $\tna{j}$ for learning $\dn$ and analyze the intended effect of the noise on the denoiser. Further details are provided in Appendix \ref{appendix:noise}.

\begin{table}[t!]
\centering
\caption{An overview of the training noise functions and the ability they are designed to induce within the denoiser. For instance, when we apply \textit{Random Deletion}, $\dn$ learns to \textit{add tokens} in order to effectively reconstruct the original sample. This is referred to as the Induced Ability in the table, and demonstrates the reason for using combinations of the presented functions. The examples given are generated after applying the respective function to the following example: \textit{Steve rode his car for 5 miles on the way home.}}
\label{tab:overview}
\resizebox{\columnwidth}{!}{
\begin{tabular}{c|c|c|c|c}
\hline
\textbf{N.} &
  \textbf{Category} &
  \textbf{Noise Function} &
  \textbf{Example} &
  \textbf{Induced Ability ($\dn$)} \\ \hline
$\tni{1}$ &
  \multirow{3}{*}{\textit{s-noise}} &
  Sentence Permutation &
  Way home Steve rode his car for 5 miles on the. &
  Permutation (for coherence) \\
$\tni{2}$ &
   &
  Random Shuffling &
  Steve for rode car his 5 miles way on home the. &
  Shuffle Groups (for coherence) \\
$\tni{3}$ &
   &
  Complete Shuffling &
  home way miles car 5 Steve rode the his on for. &
  Unscramble (for coherence) \\ \cline{1-5}
$\tni{4}$ &
  \multirow{2}{*}{\textit{c-noise}} &
  Templatization &
  Steve rode \textbf{PRON1} car for 5 miles \textbf{ADP1} the \textbf{NOUN1} home. &
  Contextual mask replacement \\
$\tni{5}$ &
   &
  Synonym Substitution &
  Steve \textbf{rides} his car \textbf{as} 5 miles on \textbf{part} way home. &
  Contextual Substitution \\ \cline{1-5}
$\tni{6}$ &
  \multirow{2}{*}{\textit{cs-noise}} &
  Random Deletion &
  Steve rode his 5 miles on home. &
  Word Addition \\
$\tni{7}$ &
   &
  Word Insertion &
  Steve \textbf{all} rode his car for \textbf{by} 5 miles on the way home. &
  Word Deletion \\
\bottomrule
\end{tabular}}
\end{table}

\subsubsection{\textit{s-noise}}
\textit{s-noise} is designed to produce variations in sentence structure through word order changes. We identify the following $3$ noising functions.

\textbf{$\tni{1}:$ Sentence Permutation:} Let the input $x$ be composed of $l$ sentences. For a sentence $s$, consisting of $k$ tokens as $\tupleb{s}{k}$, we define sentence permutation to be the selection of an index $h$ and the rotation of $s$ around $h$ to yield $\hat{s} = (s_h, s_{h+1} \dots s_k, s_1 \dots s_{h-1})$. We apply this function to a stochastically selected subset of sentences in $x$. This function is devised to teach $\dn$ how to permute sentences, as $\dn$ has to revert the permuted sample to yield the original and thus learns permutation in the process. \\
$\mathbf{\exit{1}}$: \textit{Way home Steve rode his car for 5 miles on the.}

\textbf{$\tni{2}:$ Random Shuffling:} Let the input $x$ be composed of $k$ tokens $\tupleb{x}{k}$. We randomly shuffle a contiguous subset of tokens $x_{sub} = (x_l, x_{l+1} \dots x_{l+p})$  of size $p+1$, to yield $x_{shuf} = (x_{l+i}, \dots x_{l+j})$ where $0 \leq i,j \leq p$.  We replace $x_{sub}$ with $x_{shuf}$ and repeat this procedure to generate the noised version of $x$. \\
$\mathbf{\exit{2}}$: \textit{Steve for rode car his 5 miles way on home the.}

\textbf{$\tni{3}:$ Complete Shuffling:} Unlike $\tni{2}$, we completely shuffle all the tokens in the input randomly. This function teaches $\dn$ the syntactical structure of a correct sentence as it has to learn how words chain together to form meaningful phrases to reconstruct $x$. \\
$\mathbf{\exit{3}}$: \textit{home way miles car 5 Steve rode the his on for.}

\subsubsection{\textit{c-noise}}
\textit{s-noise} is designed to teach $\dn$ to change the ordering of tokens to yield valid sentences but does not necessarily induce variations in the tokens themselves. We propose $2$ \textit{c-noise} functions that are constructed to induce contextual changes, wherein we impose additional constraints for not operating on critical entities such as numbers and units in order to preserve the solution (Appendix \ref{appendix:cnoise}).

\textbf{$\tni{4}:$ Templatization:} Masking \cite{https://doi.org/10.48550/arxiv.1810.04805} is a standard noising function where tokens are replaced by a masked token (usually [MASK]). We propose a novel templatization scheme, where instead of masking certain tokens with [MASK], we utilize their Part-Of-Speech (POS) tags along with a co-reference tracker as the mask. This provides a relatively stronger supervisory signal to $\dn$, which is necessary to maintain coherence and consistency because if multiple occurrences of one entity along with other tokens are all replaced with a singular mask, it is probable that $\dn$ might assign different entities to the masked tokens resulting in inconsistent and invalid paraphrases.\\
$\mathbf{\exit{4}}$: \textit{Steve rode \textbf{PRON1} car for 5 miles \textbf{ADP1} the \textbf{NOUN1} home.}

\textbf{$\tni{5}:$ Synonym Substitution:} We sample a subset of tokens and replace them with their synonyms (details in Appendix \ref{appendix:cnoise}).  This (like $\tni{4}$) teaches $\dn$ to generate semantically coherent replacements for multiple tokens in the input. \\
$\mathbf{\exit{5}}$: \textit{Steve \textbf{rides} his car \textbf{as} 5 miles on \textbf{part} way home.}

\subsubsection{\textit{cs-noise}}
We devise $2$ other noising functions, which can induce contextual and syntactic changes while applying the same constraints as \textit{c-noise} functions.

\textbf{$\tni{6}:$ Random Deletion:} Given the input $x = \tupleb{x}{k}$, we stochastically select and remove a subset of tokens. This serves to teach $\dn$ to add tokens when denoising. Removing one token induces the addition of one or multiple tokens, which can bring about both contextual and syntactical changes. \\ 
$\mathbf{\exit{6}}$: \textit{Steve rode his 5 miles on home.}

\textbf{$\tni{7}:$ Word Insertion:} We also add tokens to $x$ in order to teach $\dn$ deletion. If we only add random words, the denoiser would learn to delete tokens which do not agree with the context of the sample. To alleviate this, we also insert synonyms of the words present in the sample so that $\dn$ can learn to delete tokens which agree with the context of the text but are unnecessary. \\
$\mathbf{\exit{7}}$: \textit{Steve \textbf{all} rode his car for \textbf{by} 5 miles on the way home.}

\subsection{Inference Noise}
The noising functions used for training $\dn$ were generalized, as the aim was to learn a denoiser that could reverse multiple varieties of noise and produce coherent outputs. The goal of inference noising functions, however, is to \textbf{exploit the learned tendencies} of the denoiser and guide it to syntactically and contextually rich outputs which can serve as paraphrases on their own or be employed as weakly-supervised targets to learn $\p$. This is done via the fusion of grounded information with the noising functions to induce diversity while retaining solvability, as demonstrated in Figure \ref{fig:grounded}.

\subsubsection{\textit{s-noise}}
\textit{s-noise} does not lose any information (replace/delete any token $x_j^i$ in the input $x^i$) and thus has a relatively higher degree of reversibility as $\dn$ has learned to recognize the characteristics of valid sentences and if a \textit{s-noised} sample is not coherent, $\dn$ reverts it back to the original state, thus losing any diversity. To combat this, we propose \textbf{\textit{Pseudo-Adversarial Noising} (\pan)}, where we \textit{explicitly noise the input so that $\dn$ is unable to discriminate whether s-noise has been applied and thus cannot revert it} (refer to Fig \ref{fig:grounded-permute} for clarity). To accomplish this, we ground the s-noise functions on linguistic regularities (like shuffling the nodes in the constituency parse tree) along with knowledge gleaned from pretrained LMs (fluency) so that the outputs display the characteristics of valid sentences. 

\textbf{Fluency:} We model fluency as the geometric mean of the likelihood probabilities of the input sample. Specifically, we use GPT-2 \cite{radford2019language}, a pretrained auto-regressive decoder for computing the likelihood probabilities. We define the fluency of $x = \tupleb{x}{k}$ as $fl(x) = (\prod_{j=1}^{k}{p_{gpt}(x_j|x_1,x_2, \dots x_{j-1}))^{1/k}}$

\textbf{$\inni{1}:$ Sentence Permutation:} Instead of permuting at any random point as in $\tni{1}$, we discern that sentences can usually be permuted around prepositions. In the absence of a preposition within the given sentence, we permute at all possible points and select the most fluent permutation. \\
$\mathbf{\exii{1}}$: \textit{On the way home Steve rode his car for 5 miles.}

\textbf{$\inni{2}:$ Phrase Shuffling:} We identify and shuffle distinct nodes in the constituency parse tree amongst each other and choose the most fluent output. A high-level view of this function is present in Figure \ref{fig:phrase_shuffle} and further details regarding implementation are present in Appendix \ref{appendix:noise}.\\
$\mathbf{\exii{2}}$: \textit{For 5 miles Steve rode his car on the way home.}

\subsubsection{\textit{c-noise}}
\begin{figure*}[h]
    \centering
    \includegraphics[width=0.9\textwidth]{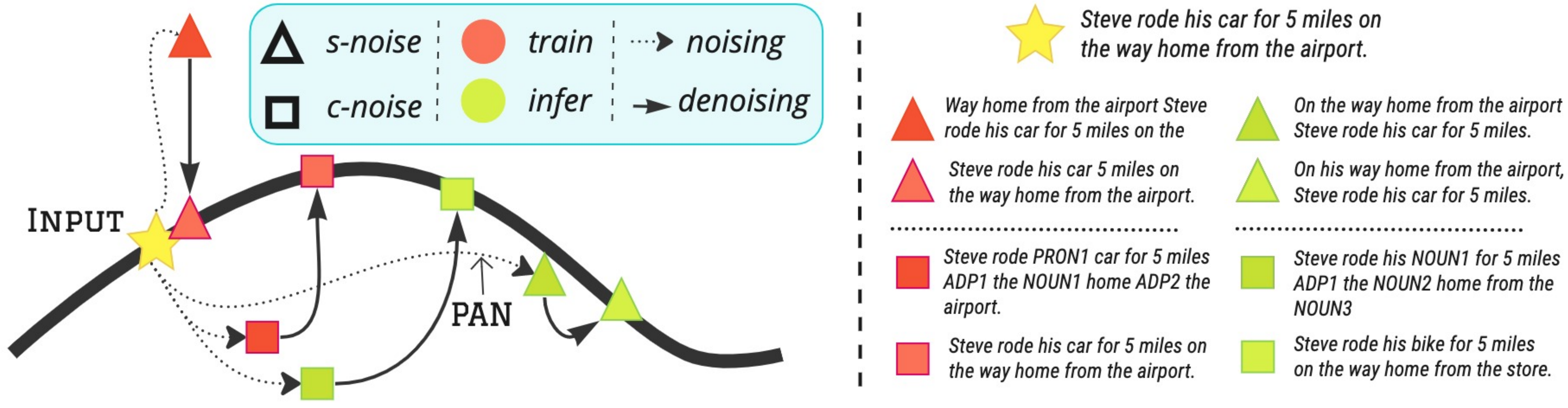}
    \caption{\textit{Left:} The data manifold corresponding to the \textit{diversity distribution of valid paraphrases} for an input sample (denoted by the star). The figure demonstrates the need for separate versions of noising functions for inference (especially for \textit{s-noise} due to its higher degree of reversibility) as the denoised sample is rendered close to the original, displaying little diversity. Using \pan locates the noised sample farther from the original and \emph{closer to the distribution of valid sentences}, ensuring that the denoiser $\dn$ is unable to discern the type of noise applied and reverse it, which happens in the training case. \textit{Right:} This is demonstrated with the help of an example. Here, the dark red and green triangles denote $\tni{1}$ and $\inni{1}$, while the dark red and green squares denote $\tni{4}$ and $\inni{3}$ respectively. In the case of $\tni{1}$, the noised sample is not grammatically sound, and the denoiser is able to implicitly identify the noise applied and reverse it (thereby losing diversity; see light red triangle). For \pan ($\inni{1}$), the noised sample is close to the data manifold, so the denoiser is unable to identify the noise applied and thereby revert it, only replacing \textit{the} with \textit{his} and adding a \textit{','}, which is syntactically diverse from the input (the star). Similarly, for templatization (denoted by the squares), masking certain entities yields richer outputs, which we exploit during inference ($\inni{3}$). We must note that individual noising functions are applied here for clarity, while multiple noising functions are applied at once in practice to generate more diverse outputs.}
    \label{fig:pan}
    \label{fig:grounded}
    \label{fig:grounded-temp}
    \label{fig:grounded-permute}
\end{figure*}

\textit{c-noise} has a relatively lower degree of reversibility as tokens get replaced with either templates or their synonyms. Thus recovery of the exact token is not always possible or desirable as similar tokens produced by the likelihood probabilities of $\dn$ are preferred. Changes in entities (usually nouns) are more optimal than changes in pronouns, articles etc., as they lead to higher-quality paraphrases (see Fig. \ref{fig:grounded-temp}). Thus, to bias $\dn$ towards induction of these changes, we ground the \textit{c-noise} functions by identifying the \textit{object} nodes in the dependency parse tree and prioritize them for templatization/substitution.

\textbf{$\inni{3}:$ Templatization:} Object nodes are given higher priority to be templatized. The effect of this grounding on $\dn$ is demonstrated in Figure \ref{fig:grounded-temp}.\\
$\mathbf{\exii{3}}$: \textit{Steve rode his \textbf{NOUN1} for 5 miles \textbf{ADP1} the \textbf{NOUN2} home.}

\textbf{$\inni{4}:$ Synonym Substitution:} Similar to $\inni{3}$, object nodes are given higher priority for substitution.\\
$\mathbf{\exii{4}}$: \textit{Steve rode his \textbf{vehicles} for 5 miles on the \textbf{path} home.}

\subsubsection{\textit{cs-noise}}
We retain the same operators with the same constraints as before, thus $\inni{5} = \tni{6}$  and $\inni{6} = \tni{7}$.

\subsection{Noise Combinations}
\label{Noise Combination}
Instead of listing all combinations utilized for training and inference, we highlight $1$ of each due to space constraints. We tabulate and analyze the complete list of the combinations used along with their intended effect and examples in Appendix \ref{appendix:combinations}. \\
\textbf{$\tna{a}:$ Random Deletion + Random Shuffling + Templatization:} $\mathbf{\exct{a}}$: \textit{his Steve rode for 5 miles the NOUN1 home.} 

\textbf{$\inna{a}:$ Sentence Permutation + Templatization:} $\mathbf{\exci{a}}$: \textit{On the way NOUN1 Steve rode his NOUN2 for 5 miles.}

\subsection{Prompting}
We utilize discrete interpretable prompts \cite{liu2021pre} for guiding the output of both $\dn$ and $\p$. For learning $\dn$, we include the control for passivization by converting a fraction of the samples (20\%) in the corpus to their passive form using a pretrained language model. Let the original sample be $x$, its passive version be $x_p$, the noised version be $\hat{x} \sim q^t(\hat{x}|x)$ and the discrete prompts be $p_{n}$ for standard denoising and $p_{pas}$ for passivization. Then, $\dn$ learns to map $[p_{pas}||\hat{x}] \mapsto x_p$, and for the remaining samples, $[p_{n}||\hat{x}] \mapsto x$ where $[.||.]$ represents text concatenation. Practically, we set $p_{pas}$ as \textit{'paraphrase passive:'} and $p_{n}$ as \textit{'paraphrase:'}. $p_{pas}$ enables $\dn$ to \textit{learn passivization as an auxiliary task}.

Our perspective of the noising search space also allows for prompting $\p$ to add a degree of control to its outputs. Specifically, we define three distinct prompts, $p_s, p_c, p_{cs}$ (as well as $p_{pas}$) for \textit{s-noise, c-noise} and \textit{cs-noise}. When we prompt $\p$ with $p_s$, we expect to induce changes in the sentence structure, while when we prompt with $p_c$, the aim is to induce changes in entities (due to the manner in which we define the noising functions). If a combination of categories is applied, we concatenate the respective prompts. For instance, if the combination $\inna{a}$ (Sec \ref{Noise Combination}) was applied, then the prompt would become $[p_s||p_c]$  This approach to prompting is a consequence of our framework design and its effects on $\p's$ outputs are shown in Figure \ref{fig:prompt}.

\subsection{\modela}
$\dn$ is parameterized by a pretrained transformer model. Specifically, we chose BART \cite{lewis2019bart} since it was pretrained with a denoising objective. Given the corpus $D^* = \{(x^1, x^1_*, p^1), \dots (x^n, x^n_*, p^n)\}$ where $p^i \in \{p_{n}, p_{pas}\}$ and $x^i_*$ is either $x$ or  $x_p$ depending upon $p^i$, we train it by minimizing the negative log-likelihood loss given as
\begin{align*}
    \mathcal{L}_{\Delta} = \mathbb{E}_{(x^i, x^i_*, p^i) \sim D^*, \hat{x^i} \sim q^t_{\pi_j}(\hat{x^i}|x^i)} \\
    [-\log{p_\Delta(x^i_*|\hat{x^i}, p^i)}]
\end{align*}

\subsection{Metrics}
\label{metrics}
We apply three metrics for automated evaluation and selection of samples to learn $\p$: similarity, numeracy and diversity. Given two documents $x^1$ and $y^1$ where $y^1$ is the generated paraphrase of $x^1$, let us denote the similarity, diversity, numeracy and the PQI (overall) score between these documents as $\sigma_{x^1,y^1}$, $\delta_{x^1,y^1}$, $\nu_{x^1,y^1}$ and $\ove{1}{1}$. The metrics similarity and numeracy together capture the correctness and solvability of $y^1$ in relation to $x^1$, while diversity captures how contextually and syntactically different the surface forms of $x^1$ and $y^1$ are. Note that since reference paraphrases are not available to us, \textit{we compute all the scores with respect to the original sample.} Further details about the implementation and design of the metrics are present in Appendix \ref{appendix:metrics}.

\subsubsection{Similarity}
To measure the semantic similarity of the documents $x^1$ and $y^1$, we use a transformer-based scorer, ParaQD \cite{https://doi.org/10.48550/arxiv.2206.08263}, which is designed for algebraic word problems and measures the similarity and solvability of the given documents.

\subsubsection{Numeracy}
We define numeracy between two documents $x^1$ and $y^1$ as the degree to which they share the same numerical entities. This metric proves especially useful because numbers are critical to the preservation of the solution for AWPs. More formally, let $N_{d_i}$ denote the list of numerical entities $(N_{d1}^{1}, N_{d1}^{2} \dots N_{d1}^{k})$ in \doc{i}. Then, we define numeracy between $x^1$ and $y^1$ as:

\begin{align*}
    \nu_{x^1, y^1} &= (\frac{|N_{x^1} \cap N_{y^1}|}{\max{(|N_{x^1}|, |N_{y^1}|)}})^p
\end{align*}

where $p$ controls the degree of penalty for hallucinating or missing a numerical entity. Note that we use a combination of $\sigma$ and $\nu$ to capture correctness, as $\nu$ alone is insufficient. 

\subsubsection{Diversity}
To compute the contextual and syntactic variations of $x^1$ and $y^1$, we use 1-BLEU \cite{papineni-etal-2002-bleu} and WPD \cite{liu-soh-2022-towards}. BLEU operates by leveraging the product of n-gram matches between $x^1$ and $y^1$ for various values of n to compute the surface similarity between the two documents; thus, we employ 1-BLEU to capture the diversity. This is especially useful in quantifying contextual changes in the entities but does not quantify syntactical shifts caused by permutation and shuffling. To tackle this, we use WPD, a metric proposed to capture the relative shift in the position of the words between $x^1$ and $y^1$. We use a weighted combination of these individual metrics to obtain an overall score for diversity as follows :
\begin{align*}
    &\dive{1}{1} = (1-BLEU_{x^1,y^1}) + WPD_{x^1, y^1}
\end{align*} 
\subsubsection{Paraphrase Quality Indicator (PQI)}
We don't utilize the weighted arithmetic mean because if the generated paraphrase $y^1$ is an exact copy of $x^1$, then using weighted arithmetic mean would yield a high score. Thus, to compute the overall PQI score $\ove{1}{1}$, we calculate the weighted arithmetic mean of the similarity, diversity and numeracy measures in the \textit{log} space. This ensures that if any of $\simi{1}{1}$, $\dive{1}{1}$ and $\nume{1}{1}$ are close to 0, then, $\ove{1}{1}$ would also be low.  More formally,
\begin{align*}
    \lambda_{x^1,y^1} = \exp(\sum_{h \in \mathcal{M}}\ln{h_{x^1,y^1}}); \mathcal{M} = \{\sigma, \delta, \nu\}
\end{align*}

\subsection{Data Filtering and Selection}

We propose a two-stage filtering approach \textbf{QF-PCF}, where we first filter out the samples using certain thresholds on the metrics to increase the quality of the selected samples (QF), and we filter them further based on the consistency of the syntactic and contextual changes in the paraphrase with the given prompt (PCF).
\subsubsection{QF: Quality Filtering}
We define $3$ thresholds $\tau_\sigma, \tau_\delta, \tau_\nu$ for $\sigma, \delta$ and $\nu$ respectively. We only consider the paraphrase $y^1$ for the next stage of filtering if it satisfies $h_{x^1, y^1} > \tau_h \forall h \in \{\sigma, \delta, \nu \}$.

\subsubsection{PCF: Prompt Consistency Filtering}
\textit{s-noise, c-noise} and \textit{cs-noise} directly correspond to the diversity metrics of WPD, (1-BLEU) and $\delta$ (overall score), respectively. Since the type of noising function has a causal relationship with the prompt, we filter out the samples where there is inconsistency between the diversity scores of the output and the prompt applied. For this, we again define $3$ thresholds $\tau_{s}, \tau_{c}, \tau_{sc}$ for the prompts $p_s, p_c, p_{cs}$. If $p_s$ is the given prompt, then $WPD \geq \tau_s$, if $p_c$ is given, then $1-BLEU \geq \tau_c$ and if $p_{cs}$ is given, then $\delta \geq \tau_{sc}$. If a combination of the prompts is given, then the intersection of the conditions is used to filter. These thresholds are set by tuning the performance on the validation set.

\subsubsection{Selection: Paraphrase-Adapted Maximum Marginal Relevance}
For the given document $q$ and a set of $n$ candidates $\{p_1, p_2, \dots p_n\}$ in bucket $C$, to select the top-$k$ candidates into the bucket $S$ ($S = \phi$ initially), inspired by MMR \cite{carbonell1998use}, we devise a new formulation, PAMMR, as follows: 
\begin{multline*}\label{eqn:selection}
argmax_{p_i \in C \setminus S} [\alpha(w_{\sigma}\sigma_{q, p_i} + w_{\delta}\delta_{q, p_i} + w_{\nu}\nu_{q, p_i}) + (1-\alpha)\min_{s_j \in S} \delta_{p_i, s_j}] \ni w_{\sigma} + w_{\delta}+ w_{\nu} = 1
\end{multline*}
where $\alpha$ is a control between the relevance and inter-sample diversity of the candidate with the already selected documents $S$. This formulation helps ensure that the selected documents are semantically similar and syntactically diverse with respect to the original $q$ but are not syntactically similar to each other.

\subsection{\modelb}
Similar to \modela, $\p$ is parameterized by BART, although any sequence generation model can be used. Given the weakly-supervised selected corpus $P = \pacop{m}$ where $p^i$ denotes the prompt, we train $\p$ by minimizing the negative log-likelihood loss given by:
\begin{align*}
    \mathcal{L}_\p = \mathbb{E}_{(x^i,y^i,p^i) \sim P}[-\log{p_\p(y^i|x^i,p^i)}]
\end{align*}

\section{Experimental Setup}
All the experiments were performed using $4$ $A100$ and $2$ $P100$ GPUs. All models, including the baselines, were trained for $15$ epochs with the initial learning rate of $8e-5$ using AdamW as the optimizer and a linear scheduler, with $10\%$ of the total steps as warm-up having a weight decay of $0.03$. Training the model took an average time of $1$ hour in the distributed setup with a combined batch size of $128$ ($32$ per GPU node). We used the base version of BART ($140M$ parameters) for initializing both \modela and \modelb. The decision to utilize a relatively smaller language model like BART ($140M$ parameters) was also influenced by deployability considerations. Further details of all hyperparameters are present in Appendix \ref{appendix:hp}.

\subsection{Datasets}
The datasets used in the experiments are: \\
$\bullet$ \textbf{AquaRAT} \cite{ling2017program} (Apache, V2.0) is an algebraic dataset consisting of $30,000$ problems in the training set for the first phase and $40,000$ questions in the second phase. There are $254$ problems for validation and $215$ problems for testing. \\
$\bullet$ \textbf{MathEM} is a proprietary dataset from our industry partner consisting of mathematics questions for students from grades $6$-$10$. There are $5000$ and $7000$ questions in the training set for the first and second phase. There are $300$ questions in the validation set and $200$ in the test set. \\
$\bullet$ \textbf{SAWP} \cite{https://doi.org/10.48550/arxiv.2206.08263} is a dataset consisting of $200$ algebraic problems (without paraphrases). We evaluate the proposed methods in a zero-shot setting (without training) on this dataset using the models trained on AquaRAT. \\
$\bullet$ \textbf{PAWP} \cite{https://doi.org/10.48550/arxiv.2206.08263} is a dataset consisting of $400$ algebraic word problems. Similar to $SAWP$, we use this dataset for zero-shot evaluation. 

\subsection{Comparative Systems}
We bifurcate the comparative systems into two distinct categories: (i) augmentation-based (ii) paraphrasers. The former utilize augmentations to generate parallel data which is leveraged to train BART to ensure consistency and fairness of comparison with our method. These approaches are not specifically designed for paraphrasing. The latter systems are explicitly designed for the task of paraphrasing. We compare and contrast \model against both categories of methods to assess its efficacy.

\subsubsection{Augmentation-based Systems} 
$\bullet$ \textbf{UDA}: We implement the unsupervised data augmentation \cite{uda} following the original settings in the paper to generate a parallel corpus. UDA leverages operations like backtranslation, TF-IDF based word replacement for augmentation. We train BART to learn the mapping from input to paraphrase. $\bullet$ \textbf{SSMBA}: We leverage the SSMBA augmentation method \cite{ng-etal-2020-ssmba} to generate parallel sentences for training. $\bullet$ \textbf{Rotom}: We use the ROTOM \cite{rotom} framework to generate paraphrases and train BART using it. During inference, we go from the original sample to paraphrase, as the results degrade if we corrupt the input.

\subsubsection{Paraphrasing systems}
$\bullet$ \textbf{SynPG}: We utilize SynPG \cite{huang-chang-2021-generating}, an unsupervised encoder-decoder based model for paraphrasing, which works by disentangling the syntactic and semantic components. $\bullet$ \textbf{UPLM}: We reproduce the unsupervised paraphrase generation using pre-trained language models \cite{uPA} with the original parameter settings. \textbf{UPSA}: In this approach \cite{liu-etal-2020-unsupervised}, simulated annealing is employed to search the space for local edits to form valid paraphrases. We follow the original hyperparameter settings.

\section{Results and Analysis}
\begin{table}[t!]
\caption{Comparative results on all datasets (columns). S: Similarity, S: Diversity, N: Numeracy, PQI: Paraphrase Quality Indicator. PQI is the weighted mean of all scores in the log space. Note that a high diversity when coupled with lower similarity and numeracy denotes that the outputs are not aligned with the input, while a high similarity and numeracy accompanied by a low diversity depicts the output is mostly copied from the input. \modela and \modelb outperform other systems comprehensively for all settings including zero-shot.}
\label{tab:results-auto}
\centering
\resizebox{\columnwidth}{!}{
\begin{tabular}{l|rrrr|rrrr|rrrr|rrrr}
\toprule
\multirow{2}{*}{\textbf{Method}} &
  \multicolumn{4}{c|}{\textbf{AquaRAT}} &
  \multicolumn{4}{c|}{\textbf{MathEM}} &
  \multicolumn{4}{c|}{\textbf{SAWP}} &
  \multicolumn{4}{c}{\textbf{PAWP}} \\ \cline{2-17} 
 &
  \multicolumn{1}{c|}{\textbf{S}} &
  \multicolumn{1}{c|}{\textbf{D}} &
  \multicolumn{1}{c|}{\textbf{N}} &
  \multicolumn{1}{c|}{\textbf{PQI}} &
  \multicolumn{1}{c|}{\textbf{S}} &
  \multicolumn{1}{c|}{\textbf{D}} &
  \multicolumn{1}{c|}{\textbf{N}} &
  \multicolumn{1}{c|}{\textbf{PQI}} &
  \multicolumn{1}{c|}{\textbf{S}} &
  \multicolumn{1}{c|}{\textbf{D}} &
  \multicolumn{1}{c|}{\textbf{N}} &
  \multicolumn{1}{c|}{\textbf{PQI}} &
  \multicolumn{1}{c|}{\textbf{S}} &
  \multicolumn{1}{c|}{\textbf{D}} &
  \multicolumn{1}{c|}{\textbf{N}} &
  \multicolumn{1}{c}{\textbf{PQI}} \\ \hline
SSMBA &
  \multicolumn{1}{r|}{0.92} &
  \multicolumn{1}{r|}{0.08} &
  \multicolumn{1}{r|}{0.9} &
  0.4 &
  \multicolumn{1}{r|}{0.93} &
  \multicolumn{1}{r|}{0.04} &
  \multicolumn{1}{r|}{0.95} &
  0.23 &
  \multicolumn{1}{r|}{0.93} &
  \multicolumn{1}{r|}{0.06} &
  \multicolumn{1}{r|}{0.97} &
  0.37 &
  \multicolumn{1}{r|}{0.91} &
  \multicolumn{1}{r|}{0.06} &
  \multicolumn{1}{r|}{0.96} &
  0.34 \\
Rotom &
  \multicolumn{1}{r|}{0.89} &
  \multicolumn{1}{r|}{0.22} &
  \multicolumn{1}{r|}{0.71} &
  0.53 &
  \multicolumn{1}{r|}{0.95} &
  \multicolumn{1}{r|}{0.08} &
  \multicolumn{1}{r|}{0.93} &
  0.34 &
  \multicolumn{1}{r|}{0.85} &
  \multicolumn{1}{r|}{0.22} &
  \multicolumn{1}{r|}{0.75} &
  0.52 &
  \multicolumn{1}{r|}{0.88} &
  \multicolumn{1}{r|}{0.22} &
  \multicolumn{1}{r|}{0.75} &
  0.53 \\
UDA &
  \multicolumn{1}{r|}{0.98} &
  \multicolumn{1}{r|}{0.03} &
  \multicolumn{1}{r|}{0.97} &
  0.22 &
  \multicolumn{1}{r|}{0.96} &
  \multicolumn{1}{r|}{0.02} &
  \multicolumn{1}{r|}{0.97} &
  0.09 &
  \multicolumn{1}{r|}{0.96} &
  \multicolumn{1}{r|}{0.03} &
  \multicolumn{1}{r|}{0.99} &
  0.19 &
  \multicolumn{1}{r|}{0.96} &
  \multicolumn{1}{r|}{0.03} &
  \multicolumn{1}{r|}{0.99} &
  0.18 \\
UPLM &
  \multicolumn{1}{r|}{0.72} &
  \multicolumn{1}{r|}{0.33} &
  \multicolumn{1}{r|}{0.66} &
  0.44 &
  \multicolumn{1}{r|}{0.67} &
  \multicolumn{1}{r|}{0.41} &
  \multicolumn{1}{r|}{0.64} &
  0.43 &
  \multicolumn{1}{r|}{0.79} &
  \multicolumn{1}{r|}{0.31} &
  \multicolumn{1}{r|}{0.76} &
  0.51 &
  \multicolumn{1}{r|}{0.71} &
  \multicolumn{1}{r|}{0.32} &
  \multicolumn{1}{r|}{0.7} &
  0.45 \\
UPSA &
  \multicolumn{1}{r|}{0.49} &
  \multicolumn{1}{r|}{0.68} &
  \multicolumn{1}{r|}{0.03} &
  0.09 &
  \multicolumn{1}{r|}{0.33} &
  \multicolumn{1}{r|}{0.59} &
  \multicolumn{1}{r|}{0.24} &
  0.09 &
  \multicolumn{1}{r|}{0.49} &
  \multicolumn{1}{r|}{0.68} &
  \multicolumn{1}{r|}{0.04} &
  0.1 &
  \multicolumn{1}{r|}{0.51} &
  \multicolumn{1}{r|}{0.67} &
  \multicolumn{1}{r|}{0.06} &
  0.13 \\
SynPG &
  \multicolumn{1}{r|}{0.55} &
  \multicolumn{1}{r|}{0.54} &
  \multicolumn{1}{r|}{0.04} &
  0.13 &
  \multicolumn{1}{r|}{0.56} &
  \multicolumn{1}{r|}{0.5} &
  \multicolumn{1}{r|}{0.21} &
  0.14 &
  \multicolumn{1}{r|}{0.43} &
  \multicolumn{1}{r|}{0.56} &
  \multicolumn{1}{r|}{0.04} &
  0.09 &
  \multicolumn{1}{r|}{0.46} &
  \multicolumn{1}{r|}{0.55} &
  \multicolumn{1}{r|}{0.04} &
  0.11 \\
\hline
\textbf{\modela} &
  \multicolumn{1}{r|}{0.98} &
  \multicolumn{1}{r|}{0.29} &
  \multicolumn{1}{r|}{0.97} &
  \textbf{0.70} &
  \multicolumn{1}{r|}{0.99} &
  \multicolumn{1}{r|}{0.36} &
  \multicolumn{1}{r|}{0.99} &
  \textbf{0.76} &
  \multicolumn{1}{r|}{0.98} &
  \multicolumn{1}{r|}{0.33} &
  \multicolumn{1}{r|}{0.99} &
  \textbf{0.74} &
  \multicolumn{1}{r|}{0.98} &
  \multicolumn{1}{r|}{0.33} &
  \multicolumn{1}{r|}{0.99} &
  \textbf{0.74} \\
\textbf{\modelb} &
  \multicolumn{1}{r|}{0.98} &
  \multicolumn{1}{r|}{0.29} &
  \multicolumn{1}{r|}{1.00} &
  \textbf{0.72} &
  \multicolumn{1}{r|}{0.98} &
  \multicolumn{1}{r|}{0.32} &
  \multicolumn{1}{r|}{0.99} &
  \textbf{0.73} &
  \multicolumn{1}{r|}{0.99} &
  \multicolumn{1}{r|}{0.3} &
  \multicolumn{1}{r|}{1.00} &
  \textbf{0.73} &
  \multicolumn{1}{r|}{0.99} &
  \multicolumn{1}{r|}{0.31} &
  \multicolumn{1}{r|}{1.00} &
  \textbf{0.74} \\
\bottomrule
\end{tabular}}
\end{table}

\subsection{Comparison with other systems}
We observe that \model outperforms the baselines by a significant margin measured by the PQI metric in Table \ref{tab:results-auto}. For instance, the percentage improvement in PQI over the best baseline on AquaRAT is \textbf{35\%}, on MathEM is \textbf{76\%} while on the zero-shot datasets it is over \textbf{35\%}. We also note that our approach's standard deviation is lower than the baselines, denoting consistency (c.f. Table \ref{tab:results})

\subsection{Performance on individual facets}
We also report the performance on individual facets like similarity, diversity and numeracy elaborated in Section \ref{metrics}. We observe that our method gives consistently high diversity while preserving similarity and numeracy. SynPG, Rotom, UPSA and UPLM also have a high diversity score, but manual examination of the outputs shows that they frequently omit information or begin hallucinating, thus lowering their similarity score. The hallucination effect is very pronounced for both SynPG and UPSA, where we observe that the \textit{numeracy} measure is extremely low, rendering the problem unsolvable. Other baselines like UDA + BART have low diversity and high similarity because they primarily copy the input.


\subsection{Human Evaluation}
\begin{table}[h]
\centering
\caption{Human evaluation results on AquaRAT for our framework and the top-$2$ baselines. We define Solvability, Diversity and Fluency as the metrics for evaluation. Mean $\pm$ std. deviation is reported. Higher is better for all metrics.}
\label{table:manual}
\begin{tabular}{c|r|r|r}
\hline
 \multirow{2}{*}{\textbf{Method}} & \multirow{2}{*}{\textbf{Solvability}}             & \multirow{2}{*}{\textbf{Diversity}} & \multirow{2}{*}{\textbf{Fluency}}\\ \\ \hline 
 \textbf{\modelb ($\p$)} & \textbf{3.59} $\pm$ 1.23 & \textbf{3.96} $\pm$ 0.73 & 3.85 $\pm$ 0.77
\\
  \textbf{\modela ($\dn$)} & 3.52 $\pm$ 1.19 & 3.86 $\pm$ 0.71 & \textbf{3.94} $\pm$ 0.78\\
  Rotom & 1.85 $\pm$ 1.54 & 1.95 $\pm$ 0.79 & 2.68 $\pm$ 0.88\\
  UPLM & 1.81 $\pm$ 1.66 & 2.27 $\pm$ 1.04 & 2.72 $\pm$ 1.31\\
 \hline
\end{tabular}
\end{table}

We manually evaluate our methods and the top-2 baselines on AquaRAT using three metrics: solvability, diversity and fluency, as shown in Table \ref{table:manual}. For the solvability metric the evaluators were requested to look for any loss of information that may render the problem unsolvable. The loss of information could include numbers, metric units or critical subject and object information. The scores are assigned in the range of 0-5 for all measures. We demonstrate that both $\dn$ and $\p$ comprehensively outperform the baselines on all metrics. We ask the evaluators\footnote{The annotators are aged between 20 and 30 with Computer Science degrees and are able to solve AWPs efficiently.} to rate each paraphrase on a scale of 0-5 for all the three measures of solvability, diversity and fluency in a blind manner. The solvability metric ensures that the solution of the paraphrase is the same as the source problem, while diversity oversees that the generated paraphrase differs in form from the original. Finally, fluency ensures that inter-sentence coherence is maintained and that the generated paraphrase is grammatically correct. For each sample, we average the responses received for that sample for all three measures mentioned above. The final reported value for each measure is the mean across samples, along with the standard deviation. We compute Cohen's Kappa ($\kappa$) to measure the inter-annotator agreement. We report average $\kappa$ values of $0.62$ for solvability, $0.56$ for diversity and $0.56$ for fluency, indicating \textit{substantial} agreement for solvability and moderate agreement for the other two measures.

From Table \ref{table:manual}, we observe that both the paraphraser and the denoiser give substantially more solvable and diverse outputs than the baselines, along with an increased degree of fluency. For solvability, we observe that the standard deviation is high for all methods but is considerably higher for the baselines (especially given the low mean). This denotes the ability of our proposed method to generate higher-quality paraphrases more frequently. However, the results indicate that there is still scope for improvement in the solvability aspect, as the aim is to generate diverse yet solvable outcomes consistently. We must note that out of the baselines (and other unsupervised methods), \textit{ours is the only method that can bring changes in the sentence structure and word order consistently.} The diversity in the baselines is usually due to hallucination or the addition/deletion of some common words, while we also perform higher order restructuring through injection of \pan along with the use of prompting and passivization. Further details are present in Appendix \ref{appendix:manual-eval}.

\subsection{Ablation Study}

\begin{table}[h]
\centering
\caption{Ablation study for both \modela and \modelb on AquaRAT. We extract out individual components in the pipeline to identify their impact and provide validation for our design choices.}
\label{table:ablations}
\begin{tabular}{l|r|r|r|r}
\hline
\textbf{Method} & \multicolumn{1}{l|}{\textbf{S}} & \multicolumn{1}{l|}{\textbf{D}} & \multicolumn{1}{l|}{\textbf{N}} & \multicolumn{1}{l}{\textbf{PQI}} \\ \hline
\textbf{\modela}  & 0.98 & 0.29 & 0.97 & \textbf{0.7}  \\ \hline
- inference noise & 0.93 & 0.30  & 0.89 & 0.65          \\ 
\textbf{\modelb}  & 0.98 & 0.29 & 1.00  & \textbf{0.72} \\ \hline
- prompts         & 0.97 & 0.24 & 0.97 & 0.67          \\
- PCF             & 0.99 & 0.28 & 0.99 & 0.71          \\
- BART + T5       & 0.97 & 0.26 & 0.98 & 0.69          \\ 
+ P2O               & 0.98 & 0.23 & 0.96 & 0.65          \\ \hline
\end{tabular}
\end{table}

We perform multiple ablations to identify the functionality of various aspects of our method empirically. We train $\p$ without prompts, without PCF, using T5 as the base model (checking for model-agnosticism), and going from the paraphrases generated by the denoiser to the original question (P2O). The results (shown in Table \ref{table:ablations}) validate our design choices. We also use training noise on the test set for $\dn$ and observe poorer results, thus justifying our definition and usage of inference noise (c.f. Appendix \ref{appendix:ablations}).

\subsection{Prompting}
\begin{figure}[h]
    \centering
    \includegraphics[width=0.5\textwidth]{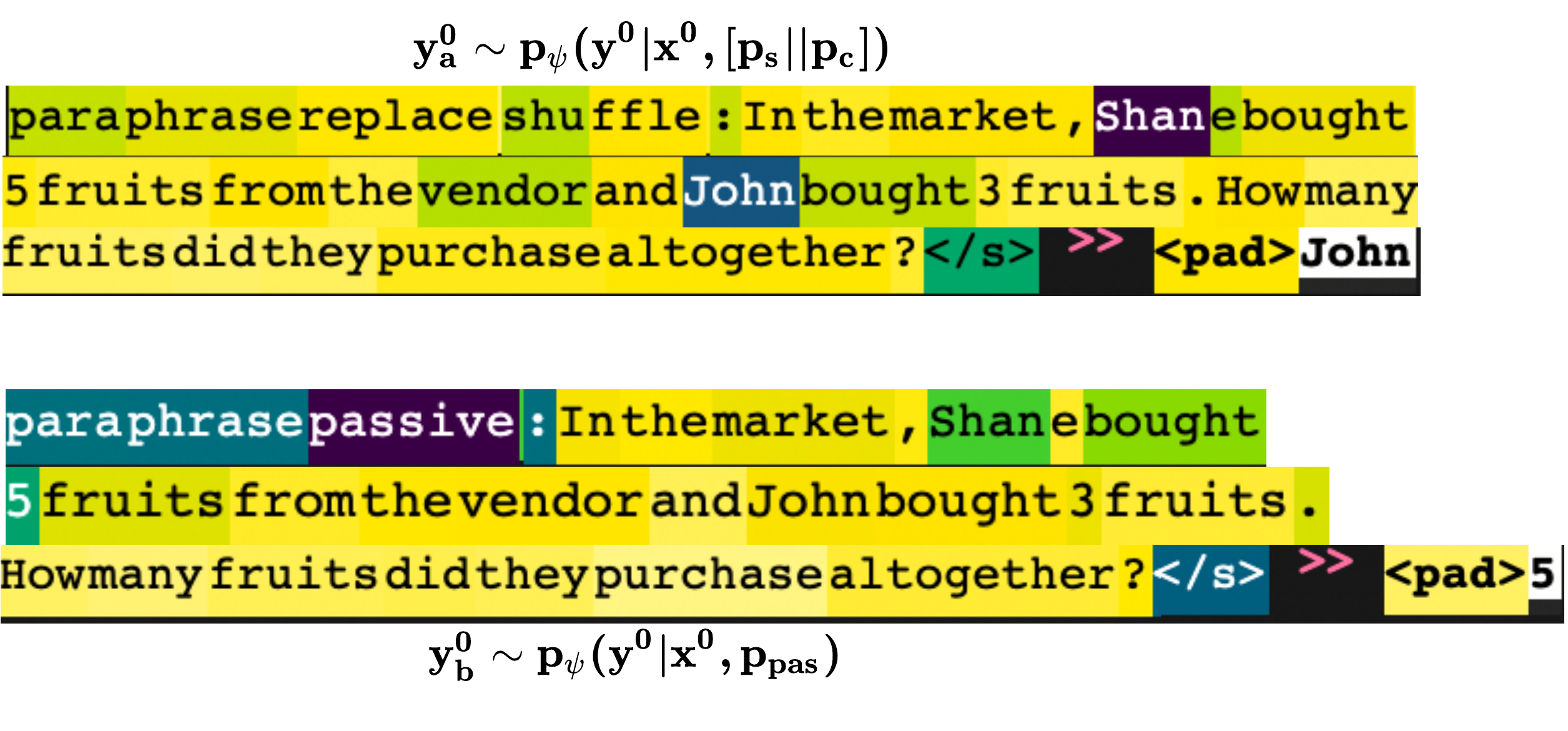}
    \caption{Feature Attribution for the same input with different prompts (\textit{paraphrase replace shuffle:} and \textit{paraphrase passive:}) given by $\p$ where darker shade denotes higher feature salience. In case of the latter prompt, a high attribution is given to the prompt word \textit{passive} denoting its importance as the model needs to restructure the output accordingly (passivization). Note the higher salience given to the first generated token (\textit{John} and \textit{5} respectively) for both prompts.}
    \label{fig:prompt-analysis}
\end{figure}

We present prompt outputs in Fig.\ref{fig:prompt} and visualize the \textbf{feature attribution} using Integrated Gradients (integral of gradients is approximated by a Riemann sum) for different prompts to increase interpretability and gain a deeper insight in Fig.\ref{fig:prompt-analysis}. In Figure \ref{fig:prompt}, when given the prompt for syntactic and contextual noise, \modelb replaces entities (\textit{fruits} with \textit{apples}) and also shuffles the structure of the input. However, when we prompt it for passivization, the model restructures the output completely and converts it into passive form. 

As per Figure \ref{fig:prompt-analysis}, note that when giving the prompt $[p_s||p_c]$, the saliency of the prompt words is relatively lower, while the named entities Shane and John are given higher weightage when generating the first token (\textit{John}). This is because the first token (in the case without passivization) could have been either name, and the sentence structure would have followed from there. However, in the case of the passive prompt, the highest saliency is given to the prompt itself (especially the word \textit{passive}). This is because, in this case, the prompt will dictate the structure of the output, as passivization requires high-level restructuring of the input. We also observe that the number \textit{5} receives a much higher weightage than the previous case, as it is the first generated token and thus requires higher attribution. This example demonstrates that changing one word in the prompt while keeping the input the same can change the internal salience assigned by the model to the input tokens and thus allow it to generate the output according to the salience assigned.
Further analysis is given in Appendix \ref{appendix:prompt}.

\subsection{Qualitative/Error Analysis} 
We present some sample outputs for both \modela and \modelb in Table \ref{tab:sample}. From the first 3 rows, we can observe how \modela is leveraging the noise to produce diverse paraphrases. The last $4$ outputs are generated by \modelb. We can observe in the $4^{th}$ and $6^{th}$ sample that the passivization prompt is being followed, but the prompt for entity changes is not being followed, as the entities remain consistent. In the last row, we can also observe that $\psi$ sometimes misunderstands entities (in this case, it considers A as the article \textit{A}). This mistake leads the model to paraphrase in a way that completely destroys the solution. This type of error is not recognized and assigned a high score by all automated metrics, but during manual evaluation, the annotators were told to rate it very low (0 or 1) in solvability. This type of error is also induced in the denoiser (infrequently) by functions such as deletion and replacement due to their higher degree of irreversibility.


\section{Conclusion}
In this paper, we proposed the novel task of paraphrasing algebraic word problems and a novel framework \textbf{\model} to achieve the same. We demonstrated our approach's ability to generate diverse paraphrases for AWPs while preserving the solution through grounded noise injection. In the future, we plan to explore the ability of our framework on general-domain corpora due to its modularity and its application as a pre-training objective for language models. We further plan on exploring \textit{interpolation of discrete prompts} for controllable text generation.

\section*{Limitations}
A limitation of our proposed method is that the combined noising combinations we use may not be the global optimum. Is there a way we can determine the best noising combination to apply, given an input sample? We plan to further explore this question by involving meta-learning techniques to estimate the optimal noising combination. Another area for improvement is controlling the surface form of the output through prompting. Our prompting method allows for this but does not always return outputs faithful to the prompt. Perhaps stricter filtering thresholds and training on a larger corpus might help with this. An exciting area to explore in more depth would be the interpolation of discrete prompts, where we could control the text in the desired manner by combining various prompts that separately signify the aspects we want to control, \textit{without necessarily training on that exact combination of prompts}. We conducted some initial experiments regarding this idea with the proposed framework, and we will explore it in detail in the future, as currently, our method does not allow for guaranteed prompt interpolation.

\section*{Ethics Statement}
The proposed solution (especially with the data we use) is intended to benefit educational systems worldwide and the NLP research community. We do not believe our method can be exploited for societal harm.

\clearpage

\bibliographystyle{unsrt}
\bibliography{references}

\clearpage

\appendix

\section{Further Noise Details}
\label{appendix:noise}

\begin{table*}[h!]
\centering
\caption{Some sample outputs from \modela and \modelb.}
\label{tab:sample}
\begin{tabular}{p{0.25\textwidth}p{0.2\textwidth}p{0.1\textwidth}p{0.25\textwidth}}
\hline
\textbf{original} &
  \textbf{noised} &
  \textbf{prompt} &
  \textbf{paraphrase} \\ \hline
A grocer makes a 25\% profit on the selling price for each bag of flour it sells. If he sells each bag for \$100 and makes \$3,000 in profit, how many bags did he sell? &
  A grocer on the selling cost for each pouch of tortillas it sold takes a 25\% profit. Because he takes \$ 3,000 in profit and sold two pouch as \$ 100, how many pouch would he sold? &
  - &
  By selling a bag of flour a grocer makes a 25\% profit. If he takes \$3,000 in profit and sells each bag for \$100, how many bags of flour would he sell? \\ \hline
A sales person gets a 10\% commission on each sale he makes. How many sales of \$250 each must he make in order to reach a salary of at least \$1000? &
  Only selling person come. gets a 10\% panel wednesday each selling he Of \$ 250 each How many selling should he come in reorder to reach a pays of at least \$ 1000? &
  - &
  A person is selling widgets. He gets a 10\% commission on each widget he sells for \$250 each. How many widgets must he sell in order to reach a price of at least \$1000? \\ \hline
A grocery sells a bag of ice for \$1.25, and makes 20\% profit. If it sells 500 bags of ice, how much total profit does it make? &
  A NOUN1 sold a NOUN2 of NOUN3 as \$ 1.25, also makes 20\% profit. If it sold 500 NOUN2 which NOUN3, how much total profit accomplishes it make? &
  - &
  A piece of candy is sold by a store for \$1.25, and a 20\% profit is made by a retailer. If 500 pieces of candy were sold by the store, how much total profit did it make? \\ \hline
While selling a watch, a shopkeeper gives a discount of 5\%. If he gives a discount of 7\%, he earns Rs. 15 less as profit. The marked price of the watch is: &
  - &
  paraphrase passive replace : &
  A discount of 5\% is given by a shopkeeper. Rs. 15 less is earned by him as profit than if he gives a discount of 7\%. The marked price of a watch is: \\ \hline
Calculate the maximum distance you can travel with \$8.50 on a cab which charges \$3.50 for the first quarter-mile and 10 cents for each additional quarter mile. &
  - &
  paraphrase fix replace shuffle : &
  A taxi charges \$3.50 for the first quarter-mile and 10 cents for each additional quarter mile. Calculate the maximum distance you can travel with \$8.50. \\ \hline
In a bag of red and green sweets, the ratio of red sweets to green sweets is 3:4. If the bag contains 120 green sweets, how many red sweets are there? &
  - &
  paraphrase passive replace : &
  Ratio is 3:4 in a bag of red and green sweets. 120 green sweets are contained in the bag, how many red sweets are there? \\ \hline
If 27 bottles of soda cost A cents, how much will B bottles cost in dollars? &
  - &
  paraphrase fix : &
  A bottle of juice costs 27 cents, how much does B bottles cost in dollars? \\ \hline
\end{tabular}
\end{table*}

\subsection{Training Noise}
\subsubsection{c-noise and cs-noise}
\label{appendix:cnoise}
For \textit{c-noise}, we note that replacing tokens with semantically related ones presents a challenge, especially in the context of word problems, as there are certain critical entities, which, if replaced or removed, would either render the question unsolvable or change the solution. To further identify key entities, we also impose the constraint of not templatizing or deleting tokens having their P.O.S. tag in (VERB, ADJECTIVE, ADVERB). This is because tokens with these tags are often critical to the solvability of AWPs. For instance, if in the sentence \textit{He gave 5 more pens}, the verb \textit{gave} was templatized/deleted and replaced by \textit{took}, the equation corresponding to the question changes. Similarly, if the adjective \textit{more} was replaced with 
\textit{less}, the solution changes. Another crucial reason for enforcing this is that the verb helps determine the object entities via the attention mechanism within the denoiser. For instance, if the verb is \textit{eating}, then the masked objects will likely be replaced with foods. If we mask the verb, there is also a chance that the verb and the objects generated might not be in logical agreement with each other. However, we allow synonym substitution and word insertion to access these tokens due to their lower destruction degree.

For $\tni{5}$, the synonyms are generated using a combination of PPDB \cite{pavlick2015ppdb}, GloVe embeddings \cite{pennington2014glove} and WordNet \cite{miller1995wordnet}. More specifically, we use the XL version of the PPDB lexical databank and 300-dimensional GloVe embeddings (6B corpus). We do this via sequentially searching PPDB, GloVe and then WordNet (using the POS tag of the token as auxiliary information).

\subsection{Inference Noise}
\subsubsection{Syntactic Noise}
We don't utilize $\tni{3}$ for inference as empirically, it leads to shuffling of numbers and other quantities amongst themselves in the denoiser outputs and thus fails to preserve the solution. 
In some cases, PAN functions may not yield coherent outputs, which are then identified and reverted by the denoiser, thus retaining semantic equivalence at the cost of diversity. 

\textbf{$\inni{1}:$ Phrase Shuffling:} For Phrase Shuffling, we use the constituency parse of the input sample. We utilize 4 linguistic regularities (designed from observation, open for addition) where we (i) preposition shuffle: insert any PP (preposition) node found as the leftmost child of its parent node (ii) conjunction shuffle: swap the nodes occurring before and after any conjunction (CC) node, (iii) verb shuffle: if multiple verbs are present, then we shuffle them (considering their dependencies) and (iv) noun-verb shuffle: we shuffle a randomly selected noun phrase (NP) with a randomly selected verb phrase (VP). We select the most fluent output from the results of these strategies.

\section{Noising Combinations}
\label{appendix:combinations}


\begin{table*}[h!]
\centering
\caption{Noise Combinations for training the denoiser, $\Delta$}
\label{tab:train-combos}
\begin{tabular}{p{0.1\textwidth}p{0.4\textwidth}p{0.4\textwidth}}
\hline
\textbf{Notation} &
  \textbf{Combination} &
  \textbf{Example} \\ \hline
$q^t_{\pi_a}$ &
  Random Deletion + Random Shuffling + Templatization &
  his Steve rode for 5 miles the NOUN1 home from the airport. \\ \hline
$q^t_{\pi_b}$ &
  Templatization &
  Steve rode PRON1 NOUN1 ADP1 5 miles ADP2 the NOUN2 home from the airport. \\ \hline
$q^t_{\pi_c}$ &
  Random Deletion + Templatization + Word Insertion &
  made Steve rode 5 miles ADP1 has DET1 home from the airport. \\ \hline
$q^t_{\pi_d}$ &
  Random Deletion + Word Insertion &
  Steve rode 5 miles on the way home from the airport. of \\ \hline
$q^t_{\pi_e}$ &
  Random Deletion + Random Shuffling + Sentence Permutation + Synonym Substitution + Templatization + Word Insertion &
  From the NOUN1 Steve rode PRON1 for car 5 on miles same way home. \\ \hline
$q^t_{\pi_f}$ &
  Random Deletion + Random Shuffling + Synonym Substitution + Word Insertion &
  Steve his ride 5 miles next household that way those home from the aerodrome. \\ \hline
$q^t_{\pi_g}$ &
  Random Deletion + Synonym Substitution + Templatization + Word Insertion &
  Steve rode ADP1 5 miles ADP2 the NOUN1 residence from the airport. \\ \hline
$q^t_{\pi_h}$ &
  Random Deletion + Sentence Permutation + Synonym Substitution + Word Insertion &
  On this home from airport Steve rode up he car for 5 by miles. \\ \hline
$q^t_{\pi_i}$ &
  Complete Shuffling + Synonym Substitution &
  for manner car rode the 5 from airport residence on. my Steve miles which \\ \hline
$q^t_{\pi_j}$ &
  Complete Shuffling + Random Deletion + Word Insertion &
  the miles same rode 5. home airport Steve rides from on path car \\ \hline
$q^t_{\pi_j}$ &
  Complete Shuffling + Random Deletion + Word Insertion &
  the miles same rode 5. home airport Steve rides from on path car \\ \hline
\end{tabular}
\end{table*}

\begin{table*}[h!]
\caption{Noising Combinations (and the associated prompts) for inference}
\label{tab:inf-combos}
\begin{tabular}{p{0.1\textwidth}p{0.3\textwidth}p{0.35\textwidth}p{0.2\textwidth}}
\hline
\textbf{Notation} &
  \textbf{Combination} &
  \textbf{Example} &
  \textbf{Prompt} \\ \hline
$q^i_{\pi_a}$ &
  Sentence Permutation + Templatization &
  From the NOUN1 Steve rode his NOUN2 for 5 miles on the way home. &
  paraphrase replace shuffle : \\ \hline
$q^i_{\pi_b}$ &
  Phrase Shuffling + Synonym Substitution &
  Steve for 5 miles ride his automobile on of manner home from the aerodrome. &
  paraphrase replace shuffle : \\ \hline
$q^i_{\pi_c}$ &
  Phrase Shuffling + Templatization &
  Steve for 5 miles rode his NOUN1 on the NOUN2 home from the airport. &
  paraphrase replace shuffle : \\ \hline
$q^i_{\pi_d}$ &
  Sentence Permutation/Phrase Shuffle + Synonym Substitution &
  Steve as 5 miles rode her vehicles on the manner home from the aerodrome. &
  paraphrase replace shuffle : \\ \hline
$q^i_{\pi_e}$ &
  Synonym Substitution + Templatization &
  Steve riding his NOUN1 for 5 miles on the NOUN2 home from the aerodrome. &
  paraphrase replace : \\ \hline
$q^i_{\pi_f}$ &
  Sentence Permutation + Synonym Substitution &
  On the path home from the aerodrome Steve rode his automobile for 5 miles. &
  paraphrase replace shuffle : \\ \hline
$q^i_{\pi_g}$ &
  Random Deletion + Templatization + Word Insertion &
  horseback Steve rode his NOUN1 ADP1 which 5 miles the NOUN2 from the airport. &
  paraphrase fix replace : \\ \hline
$q^i_{\pi_h}$ &
  Phrase Shuffling + Sentence Permutation + Synonym Substitution &
  way the On household from the airport rode vehicles Steve his for 5 miles. &
  paraphrase replace shuffle : \\ \hline
$q^i_{\pi_i}$ &
  Random Deletion + Word Insertion &
  Steve rode his 5 on miles of the home from the airport. &
  paraphrase fix : \\ \hline
$q^i_{\pi_j}$ &
  Sentence Permutation/Phrase Shuffle + Synonym Substitution + Templatization &
  Steve for 5 miles rode his NOUN1 on DET1 path home from the aerodrome. &
  paraphrase replace shuffle : \\ \hline
\end{tabular}
\end{table*}

The noising combinations used for training and inference are given in Table \ref{tab:train-combos} and Table \ref{tab:inf-combos} for AquaRAT. We combine multiple types of noise during training $\dn$ to generalize its denoising ability to a greater extent. The combination usually requires the model to deal with both syntactical and contextual noise. The combinations are also designed in a manner to induce particular changes. For instance, the combination of addition and deletion teaches the denoiser to add and delete or replace words; similarly, the contextual noising functions are also designed to teach the model mainly about replacing tokens. We extensively utilize deletion, addition, templatization and synonym substitution to teach the model to make these changes. At the same time, we don't heavily focus on the syntactic functions during training as $\dn$ might learn to reverse even the grounded \textit{PAN} functions. Deletion and addition (along with replacement) are also critical because they teach the denoiser proper sentence structure, as it can learn to delete or add tokens to make the outcome a valid sentence.

During inference, we usually sample the noise combinations by combining one syntactic noising operator and one contextual noising operator as we want to induce both changes in the outputs. We utilize the syntactic+contextual functions, grouping both of them within noise combinations, although their usage is not as heavy as within the training combinations due to their potential destructive properties. The outputs of the noise combinations are also given in Table \ref{tab:inf-combos}. We also present the discrete value of the prompts we use for the combinations, determined by the noise category applied. Note that the combinations can be changed based on the dataset and the structural type of outputs desired.

        

\begin{figure}[h]
    \centering
    \includegraphics[width=0.5\textwidth]{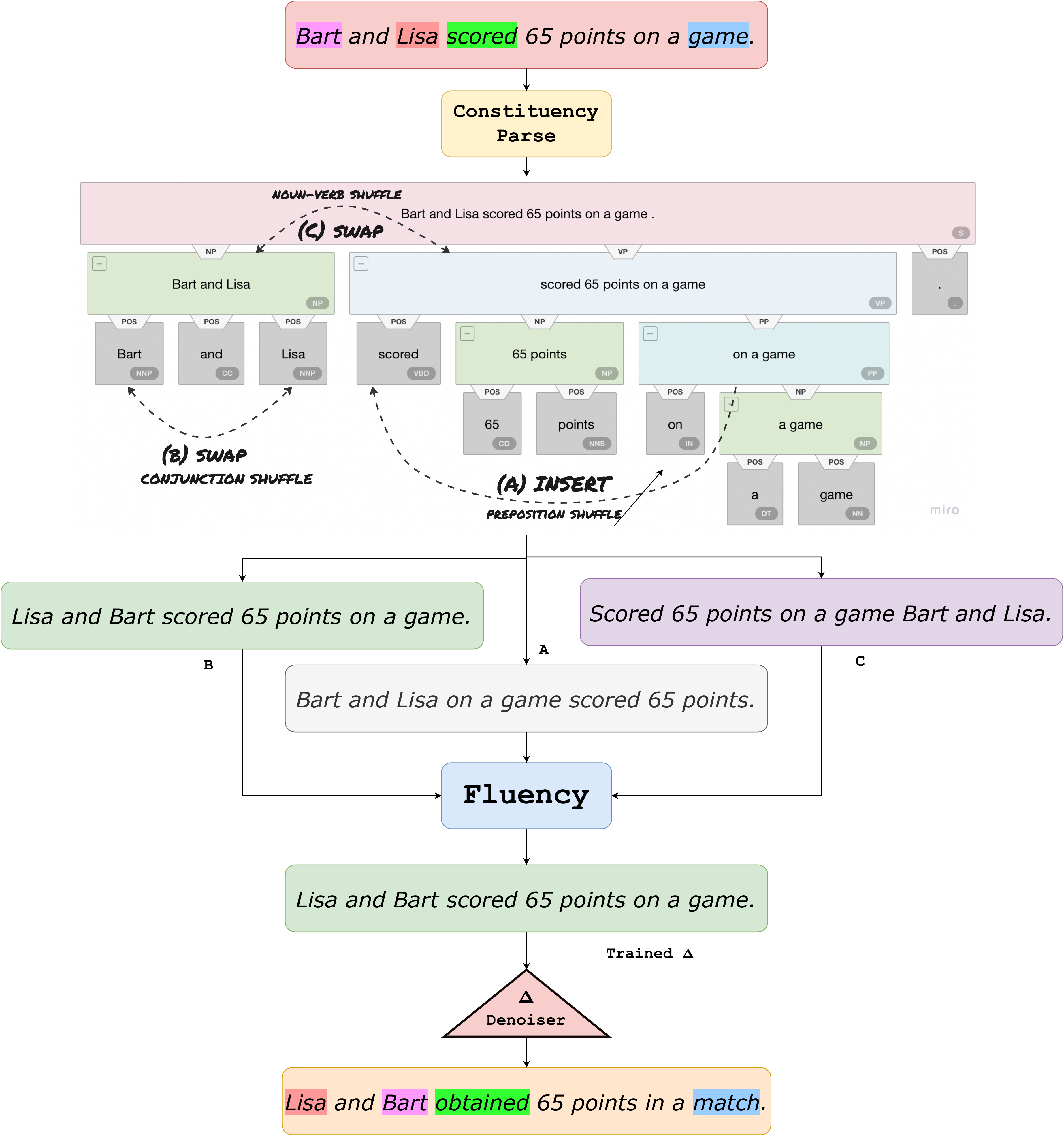}
    \caption{An overview of how $\inni{2}$, Phrase Shuffling, uses grounded information through shuffling the nodes in the constituency parse tree of the input, combined with likelihood probabilities of a pretrained LM in the form of fluency to exploit $\dn$. The output of $\dn$ serves as a paraphrase in itself, and is also used to train $\p$. The constituency parse output is from \href{https://demo.allennlp.org/constituency-parsing}{AllenNLP}.}
    \label{fig:phrase_shuffle}
\end{figure}

\begin{table*}[h!]
\centering
\caption{Automated evaluation across all methods (mean $\pm$ std. deviation).}
\begin{tabular}{clrrrr}
\hline
\multicolumn{1}{l}{\multirow{2}{*}{\textit{\textbf{Dataset}}}} & \multirow{2}{*}{\textbf{Method}} & \multirow{2}{*}{\textbf{Similarity}}             & \multirow{2}{*}{\textbf{Diversity}} & \multirow{2}{*}{\textbf{Numeracy}} & \multirow{2}{*}{\textbf{PQI}} \\ \\ \hline 

\multirow{6}{*}{AquaRAT}                                       & SSMBA + BART                       & 0.92 $\pm$ 0.27  & 0.08 $\pm$ 0.05 & 0.9 $\pm$ 0.23 & 0.4 $\pm$ 0.21 \\

& Rotom + BART                       & 0.89 $\pm$ 0.32  & 0.22 $\pm$ 0.06 & 0.71 $\pm$ 0.35 & 0.53 $\pm$ 0.21 \\
& UDA + BART & 0.98 $\pm$ 0.15 & 0.03 $\pm$ 0.03 & 0.97 $\pm$ 0.13 & 0.22 $\pm$ 0.23 \\

& UPLM & 0.72 $\pm$ 0.45 & 0.33 $\pm$ 0.17 & 0.66 $\pm$ 0.38 & 0.44 $\pm$ 0.29 \\
& UPSA & 0.49 $\pm$ 0.28 & 0.68 $\pm$ 0.06 & 0.03 $\pm$ 0.12 & 0.09 $\pm$ 0.16 \\
& SynPG & 0.55 $\pm$ 0.41 & 0.54 $\pm$ 0.07 & 0.04 $\pm$ 0.07 & 0.13 $\pm$ 0.19 \\
& \textbf{Denoiser (ours)} & 0.98 $\pm$ 0.14 & 0.29 $\pm$ 0.09 & 0.97 $\pm$ 0.13 & \textbf{0.70} $\pm$ 0.11 \\
& \textbf{Paraphraser (ours)} & 0.98 $\pm$ 0.14 & 0.29 $\pm$ 0.09 & 1.00 $\pm$ 0.04 & \textbf{0.72} $\pm$ 0.11

\\ \hline

\multirow{6}{*}{MathEM}                                       & SSMBA + BART                       & 0.93 $\pm$ 0.26  & 0.04 $\pm$ 0.06 & 0.95 $\pm$ 0.21 & 0.23 $\pm$ 0.25 \\

& Rotom + BART                       & 0.95 $\pm$ 0.22  & 0.08 $\pm$ 0.08 & 0.93 $\pm$ 0.21 & 0.34 $\pm$ 0.27 \\
& UDA + BART & 0.96 $\pm$ 0.21 & 0.02 $\pm$ 0.04 & 0.97 $\pm$ 0.15 & 0.09 $\pm$ 0.17 \\

& UPLM & 0.67 $\pm$ 0.44 & 0.41 $\pm$ 0.20 & 0.64 $\pm$ 0.43 & 0.43 $\pm$ 0.33 \\

& UPSA & 0.33 $\pm$ 0.35 & 0.59 $\pm$ 0.11 & 0.24 $\pm$ 0.41 & 0.09 $\pm$ 0.19 \\

& SynPG & 0.56 $\pm$ 0.39 & 0.50 $\pm$ 0.11 & 0.21 $\pm$ 0.40 & 0.14 $\pm$ 0.26 \\
& \textbf{Denoiser (ours)} & 0.99 $\pm$ 0.08 & 0.36 $\pm$ 0.12 & 0.99 $\pm$ 0.07 & \textbf{0.76} $\pm$ 0.08 \\
& \textbf{Paraphraser (ours)} & 0.98 $\pm$ 0.08 & 0.32 $\pm$ 0.11 & 0.99 $\pm$ 0.08 & \textbf{0.73} $\pm$ 0.08

\\ \hline

\multirow{6}{*}{SAWP}                                       & SSMBA + BART                       & 0.93 $\pm$ 0.25  & 0.06 $\pm$ 0.05 & 0.97 $\pm$ 0.13 & 0.37 $\pm$ 0.23 \\

& Rotom + BART                       & 0.85 $\pm$ 0.35  & 0.22 $\pm$ 0.06 & 0.75 $\pm$ 0.33 & 0.52 $\pm$ 0.23 \\
& UDA + BART & 0.96 $\pm$ 0.20 & 0.03 $\pm$ 0.04 & 0.99 $\pm$ 0.10 & 0.19 $\pm$ 0.23 \\

& UPLM & 0.79 $\pm$ 0.40 & 0.31 $\pm$ 0.11 & 0.76 $\pm$ 0.35 & 0.51 $\pm$ 0.28 \\

& UPSA & 0.49 $\pm$ 0.27 & 0.68 $\pm$ 0.06 & 0.04 $\pm$ 0.11 & 0.10 $\pm$ 0.17 \\

& SynPG & 0.43 $\pm$ 0.40 & 0.56 $\pm$ 0.06 & 0.04 $\pm$ 0.12 & 0.09 $\pm$ 0.17 \\

& \textbf{Denoiser (ours)} & 0.98 $\pm$ 0.12 & 0.33 $\pm$ 0.09 & 0.99 $\pm$ 0.09 & \textbf{0.74} $\pm$ 0.11 \\

& \textbf{Paraphraser (ours)} & 0.99 $\pm$ 0.07 & 0.30 $\pm$ 0.09 & 1.00 $\pm$ 0.04 & \textbf{0.73} $\pm$ 0.08 

\\ \hline
\multirow{6}{*}{PAWP}                                       & SSMBA + BART                       & 0.91 $\pm$ 0.28  & 0.06 $\pm$ 0.05 & 0.96 $\pm$ 0.16 & 0.34 $\pm$ 0.24 \\

& Rotom + BART                       & 0.88 $\pm$ 0.33  & 0.22 $\pm$ 0.06 & 0.75 $\pm$ 0.34 & 0.53 $\pm$ 0.22 \\
& UDA + BART & 0.96 $\pm$ 0.21 & 0.03 $\pm$ 0.04 & 0.99 $\pm$ 0.09 & 0.18 $\pm$ 0.23 \\

& UPLM & 0.71 $\pm$ 0.45 & 0.32 $\pm$ 0.13 & 0.70 $\pm$ 0.39 & 0.45 $\pm$ 0.30 \\

& UPSA & 0.51 $\pm$ 0.27 & 0.67 $\pm$ 0.06 & 0.06 $\pm$ 0.16 & 0.13 $\pm$ 0.19 \\

& SynPG & 0.46 $\pm$ 0.40 & 0.55 $\pm$ 0.06 & 0.04 $\pm$ 0.10 & 0.11 $\pm$ 0.18 \\

& \textbf{Denoiser (ours)} & 0.98 $\pm$ 0.13 & 0.33 $\pm$ 0.09 & 0.99 $\pm$ 0.09 & \textbf{0.74} $\pm$ 0.11\\

& \textbf{Paraphraser (ours)} & 0.99 $\pm$ 0.05 & 0.31 $\pm$ 0.08 & 1.00 $\pm$ 0.00 & \textbf{0.74} $\pm$ 0.06\\

\hline

\end{tabular}

\label{tab:results}
\end{table*}

\begin{table*}[h!]
\centering
\caption{(1-BLEU) and WPD measures for overall diversity measure.}
\begin{tabular}{clrr}
\hline
\multicolumn{1}{l}{\multirow{2}{*}{\textit{\textbf{Dataset}}}} & \multirow{2}{*}{\textbf{Method}} & \multirow{2}{*}{\textbf{1-BLEU}}             & \multirow{2}{*}{\textbf{WPD}}  \\ \\ \hline 

\multirow{6}{*}{AquaRAT} & SSMBA + BART                       & 0.12 $\pm$ 0.08 & 0.01 $\pm$ 0.02 \\

& Rotom + BART                       & 0.32 $\pm$ 0.09  & 0.07 $\pm$ 0.03 \\
& UDA + BART & 0.04 $\pm$ 0.05 & 0.00 $\pm$ 0.01  \\

& UPLM & 0.47 $\pm$ 0.20 & 0.11 $\pm$ 0.16  \\
& UPSA & 0.96 $\pm$ 0.06 & 0.68 $\pm$ 0.06 \\
& SynPG & 0.77 $\pm$ 0.10 & 0.19 $\pm$ 0.06 \\
& \textbf{Denoiser (ours)} & 0.41 $\pm$ 0.13 & 0.11 $\pm$ 0.08  \\
& \textbf{Paraphraser (ours)} & 0.40 $\pm$ 0.12 & 0.13 $\pm$ 0.09
\\ \hline

\multirow{6}{*}{MathEM} & SSMBA + BART                       & 0.07 $\pm$ 0.09 & 0.01 $\pm$ 0.02 \\

& Rotom + BART                       & 0.12 $\pm$ 0.12  & 0.03 $\pm$ 0.03 \\
& UDA + BART & 0.02 $\pm$ 0.06 & 0.00 $\pm$ 0.01  \\

& UPLM & 0.58 $\pm$ 0.24 & 0.17 $\pm$ 0.17  \\
& UPSA & 0.86 $\pm$ 0.13 & 0.19 $\pm$ 0.09 \\
& SynPG & 0.72 $\pm$ 0.15 & 0.18 $\pm$ 0.09\\

& \textbf{Denoiser (ours)} & 0.47 $\pm$ 0.16 & 0.20 $\pm$ 0.11  \\
& \textbf{Paraphraser (ours)} & 0.43 $\pm$ 0.15 & 0.15 $\pm$ 0.12
\\ \hline

\multirow{6}{*}{SAWP} & SSMBA + BART                       & 0.10 $\pm$ 0.08 & 0.01 $\pm$ 0.01 \\

& Rotom + BART                       & 0.31 $\pm$ 0.09  & 0.07 $\pm$ 0.04 \\
& UDA + BART & 0.04 $\pm$ 0.06 & 0.01 $\pm$ 0.01  \\

& UPLM & 0.46 $\pm$ 0.16 & 0.08 $\pm$ 0.05  \\
& UPSA & 0.95 $\pm$ 0.05 & 0.27 $\pm$ 0.09 \\
& SynPG & 0.80 $\pm$ 0.08 & 0.18 $\pm$ 0.07\\
& \textbf{Denoiser (ours)} & 0.49 $\pm$ 0.13 & 0.08 $\pm$ 0.04  \\
& \textbf{Paraphraser (ours)} & 0.44 $\pm$ 0.14 & 0.09 $\pm$ 0.05
\\ \hline

\multirow{6}{*}{PAWP} & SSMBA + BART                       & 0.09 $\pm$ 0.08 & 0.01 $\pm$ 0.01 \\

& Rotom + BART                       & 0.31 $\pm$ 0.09  & 0.08 $\pm$ 0.04 \\
& UDA + BART & 0.04 $\pm$ 0.07 & 0.00 $\pm$ 0.01  \\

& UPLM & 0.47 $\pm$ 0.17 & 0.09 $\pm$ 0.09  \\
& UPSA & 0.94 $\pm$ 0.06 & 0.26 $\pm$ 0.08 \\
& SynPG & 0.79 $\pm$ 0.08 & 0.18 $\pm$ 0.06\\
& \textbf{Denoiser (ours)} & 0.49 $\pm$ 0.14 & 0.09 $\pm$ 0.05  \\
& \textbf{Paraphraser (ours)} & 0.47 $\pm$ 0.17 & 0.09 $\pm$ 0.09
\\ \hline
\end{tabular}
\end{table*}



 
  
  

\section{Prompting: Analysis}
\label{appendix:prompt}
We can see the effect of prompting in Figure \ref{fig:prompt}. When given the prompt for syntactic and contextual noise, $\p$ replaces entities (\textit{fruits} with \textit{apples}) and also shuffles the structure of the input. However, when we prompt it with the prompt for passivization, the model restructures the output completely and converts it into passive form. 

We visualized the feature attribution when generating the first token in Figure \ref{fig:prompt-analysis} to gain a deeper insight into this phenomenon. We do this via Integrated Gradients \cite{sundararajan2017axiomatic}, where the integral of gradients is approximated with respect to the inputs along a particular path using a Riemann Sum. We use Ecco \cite{alammar-2021-ecco} to visualise the feature attribution. We can observe in the figure that when giving the prompt $[p_s||p_c]$, the saliency of the prompt words is relatively lower, while the named entities Shane and John are given higher weightage when generating the first token (\textit{John}). This is because the first token (in the case without passivization) could have been either name, and the sentence structure would have followed from there. 

However, in the case of the passive prompt, the highest saliency is given to the prompt itself (especially the word \textit{passive}). This is because, in this case, the prompt will dictate the structure of the output, as passivization requires high-level restructuring of the input. We also observe that the number \textit{5} receives a much higher weightage than the previous case, as it is the first generated token and thus requires higher attribution. This example demonstrates that changing one word in the prompt while keeping the input the same can change the internal salience assigned by the model to the input tokens and thus allow it to generate the output according to the salience assigned.

\section{Metrics}
\label{appendix:metrics}

Note that since reference paraphrases are not available to us, we compute all the scores with respect to the original.

\subsection{Similarity}
We don't utilize the standard overlap measures like BLEU and ROUGE scores due to their inability to detect the correctness of the paraphrase (preservation of critical information like numbers and units) and properly score the contextual change property of AWPs. We calculate the cosine similarity of the embeddings to measure the final score and linearly scale it between 0 and 1. Given the cosine similarity of $x^i$ and $y^i$ as $\cos{(x^i, y^i)}$, $\sigma_{x^i, y^i}$ is defined as:
\begin{align*}
    \simi{i}{i} = \frac{\cos{(x^i, y^i)} + 1}{2}
\end{align*}

\subsection{Numeracy}
We set $p$ as a control measure for hallucination or emission of numerical entities. The higher the value of $p$, the harsher the penalty as the numeracy score decreases exponentially. We set $p=3$, which is a relatively harsher penalty to encourage the selection of high-quality samples.

\subsection{PQI}
We don't utilize the weighted arithmetic mean because if the generated paraphrase $y^1$ is an exact copy of $x^1$, then using weighted arithmetic mean would yield a score of $w_{\sigma} + w_{\nu}$ (as $\simi{1}{1}=1, \dive{1}{1}=0$ and $\nume{1}{1}=1$), which would be a relatively high score for a poor paraphrase. Thus, to increase the overall penalty for a poor performance in any of the facets of similarity, diversity and numeracy, we utilize weighted geometric mean, which is nothing but the weighted arithmetic mean of the scores in the \textit{log space}. This ensures that if any of the scores are low, the PQI score is also low, thus becoming reflective of the quality of the paraphrase.

\section{Experimental Setup}
\label{appendix:hp}

For training the model, we did not perform extensive hyperparameter tuning. Our core method is not dependent upon the model architecture but rather the design of our noising functions, framework and prompting. For noising, we used the spaCy library combined with benepar \cite{kitaev-klein-2018-constituency} for constituency parsing and spaCy for dependency parsing. We utilize diverse beam search \cite{vijayakumar2018diverse}, with diversity penalty set to 10.0, the number of beams set to 6 and the number of beam groups set to 3. We set the random seed for all experiments to 3407 and load the initial pretrained model weights from HuggingFace \cite{wolf-etal-2020-transformers}. 

\subsection{Prompting}
We set $p_s$ to \textit{shuffle}, $p_c$ to \textit{replace} and $p_{sc}$ to \textit{fix} to make it more interpretable and give $\p$ a stronger signal about what is required in the output. Other variations can also be used for these prompts. For passivization, we utilize Styleformer (\href{https://github.com/PrithivirajDamodaran/Styleformer}{reference}) and pass the outputs of this model through a grammar corrector to make the passive form grammatically correct.

\subsection{Diversity}
We set the value of $v_1$ and $v_2$ to 0.6 and 0.4, respectively, giving a slightly higher weightage to BLEU  diversity. This is done because BLEU-Diversity can capture word order changes (to a minor extent as this is its secondary purpose) through n-gram overlap. At the same time, WPD is not as effective in detecting entity changes (its secondary purpose). We report the BLEU-Diversity and WPD scores in Table 6. $\dn$ and $\p$ have relatively higher scores compared to augmentation-based generation approaches like UDA and SSMBA. SynPG and UPSA have higher diversity scores than our method due to hallucination because, from the other metrics (similarity and numeracy) reported in Table 1 and Table 5, we observe that they do not produce semantically meaningful paraphrases.

\subsection{Parameters}
We set $w_\sigma$, $w_\delta$ and $w_\nu$ to $0.5, 0.25, and 0.25$, respectively, assigning a higher score to the similarity measure as the correctness of the output is the most important aspect in our task. We set $\tau_\sigma, \tau_\delta, \tau_\nu$ to $0.9, 0.15, 0.6$ after observing the mean scores on the validation set and applying a slightly harsher threshold for higher quality selection. Observing the mean scores for BLEU, WPD and overall diversity on the validation set, we set the value of $\tau_s, \tau_c, \tau_{s+c}$ as 0.1, 0.3 and 0.25.
We set $\alpha$ to 0.65 to maintain control between the relevance of the selected documents and their diversity with the other selected documents. If we set $\alpha$ close to 1, then we would ignore inter-sample diversity, which can lead to the selection of documents having similar surface forms. Lower values can lead to the selection of irrelevant documents. Thus, we set the value as 0.65, assigning a higher weightage to relevance but not completely ignoring the inter-sample diversity factor. We set $k=2$ for training \modelb and follow this setting for the baselines (selecting $2$ outputs per input).

\section{Manual Evaluation}
\label{appendix:manual-eval}

We perform manual evaluation on the AquaRAT dataset (we don't manually evaluate other datasets due to resource constraints) apart from the automated metrics reported earlier. We generate paraphrases using our denoiser ($\dn$), paraphraser ($\p$), and the two best baselines: Rotom and UPLM. We provide the four sets to two annotators pursuing Computer Science degrees who are adept at solving Algebraic Word Problems. The annotators were fairly compensated for the task. They were asked to rate each paraphrase on a scale of 0-5 for all the three measures of solvability, diversity and fluency in a blind manner. The solvability metric ensures that the solution of the paraphrase is the same as the source problem and that the generated paraphrase is comprehensible for the learner. The annotators are asked to give a high rating of 4 or 5 only if it satisfies the mentioned conditions.
Similarly, for diversity, the annotators are asked to provide a high rating only if the generated paraphrase differs in form from the original. Finally, fluency ensures that inter-sentence coherence is maintained and that the generated paraphrase is grammatically correct. A high rating of 4 or 5 is applicable only if the above conditions are satisfied. A medium-level rating of 2 or 3 is applicable if the annotators think the problem is still understandable and approachable. While a low rating of 0 or 1 is applicable when the problem is completely unsolvable or grammatically incorrect.

We report the three metrics in Table \ref{table:manual}. For each sample, we average the responses received for that sample for all three measures mentioned above. The final reported value for each measure is the mean across samples, along with the standard deviation. We compute Cohen's Kappa ($\kappa$) to measure the inter-annotator agreement. We report average $\kappa$ values of 0.62 for solvability, 0.56 for diversity and 0.56 for fluency, indicating \textit{substantial} agreement for solvability and moderate agreement for the other two measures.

From Table \ref{table:manual}, we observe that both the paraphraser and the denoiser give substantially more solvable and diverse outputs than the baselines, along with an increased degree of fluency. For solvability, we observe that the standard deviation is high for all methods but is considerably higher for the baselines (especially given the low mean). This denotes the ability of our proposed method to generate higher-quality paraphrases more frequently. However, the results indicate that there is still scope for improvement in the solvability aspect, as the aim is to generate diverse yet solvable outcomes consistently. 

For diversity, the generation of completely unrelated outputs would have achieved a high score (around 1) on the automated evaluation metrics, but we instructed the annotators to assign it a low score (0) for diversity, as there is no relation to the input. This is especially observed in the case of UPLM, where the automated diversity score is higher but much lower in manual evaluation. Also, in the case of diversity, we must note that due to the multi-sentence structure of AWPs, if one out of 2 sentences gets shuffled, WPD assigns a low score (around 0.1 or 0.2) due to its inherent design. However, human annotators assign it a higher score ($\geq$ 3 out of 5), which is reflected in our diversity scores. Finally, we must note that out of the baselines (and other unsupervised methods), \textbf{ours is perhaps the only method that can bring changes in the sentence structure and word order consistently.} The diversity in the baselines is usually due to hallucination or the addition/deletion of some common words, while we also perform higher order restructuring through injection of \textit{pseudo-adversarial noise} along with the use of prompting and passivization.

\section{Ablation Analysis}
\label{appendix:ablations}
We perform several ablations of \textit{Paraphraser} ($\p$) as shown in Table \ref{table:ablations}. We train a version without prompts and without PCF-based filtering explained in Section 3.8.2. We note that prompts, in general, help in producing highly diverse outputs and also ensure consistency of solution to the original problem by preserving key quantities like numbers as demonstrated by the \textit{Numeracy} metric. Training without PCF does decrease diversity, though it is not very pronounced. However, on manual examination of the data, the version of the paraphraser trained on data obtained with PCF-based filtering is more coherent and faithful to the prompts provided. We also perform an ablation where we parameterized $\p$ with T5. This leads to a drop in all three metrics, even though T5 has more parameters. This is maybe owing to the fact that BART, which is pre-trained on a denoising objective, is more suited to our task than T5. Another ablation performed for the paraphraser is P2O, where we reverse the inputs and outputs for training $\p$. The model learns to reproduce the original question from the paraphrases generated from $\dn$. We observe that the model trained in this fashion performs relatively poorly regarding diversity and numeracy. This demonstrates that $\dn$ generates good paraphrases that can act as training data to train other generative models. 

Finally, we perform an ablation on $\dn$, where we replace the inference noising functions with training noise functions and calculate the scores for the generated paraphrases on the test set. It yields a considerable drop in numeracy and similarity metrics, indicating the model generates paraphrases where key quantities like numbers may not be preserved. This is most likely (also confirmed with manual examination) due to the complete shuffling function, which aids in producing diverse outputs (hence the high diversity score), but key aspects like numbers, units, and key-phrases may not be preserved. This ablation supports our hypothesis for injection of \textit{pseudo-adversarial} grounded noise and our removal of complete shuffling from the inference noising functions.

\end{document}